\documentclass[10pt]{article} 
\usepackage[accepted]{rlc}

\usepackage{amssymb}            
\usepackage{mathtools}          
\usepackage{mathrsfs}           
\mathtoolsset{showonlyrefs}     
\usepackage{graphicx}           
\usepackage{subcaption}         
\usepackage[space]{grffile}     
\usepackage{url}                

\usepackage[utf8]{inputenc} 
\usepackage[T1]{fontenc}    
\usepackage{hyperref}       
\usepackage{booktabs}       
\usepackage{amsfonts}       
\usepackage{nicefrac}       
\usepackage{microtype}      
\usepackage{xcolor}         
\usepackage{float}
\restylefloat{table}

\usepackage{caption}
\usepackage{booktabs}
\usepackage{wrapfig}
\usepackage{multirow}
\usepackage{bm}

\begin{document}
\title{Can Differentiable Decision Trees Enable Interpretable Reward Learning from Human Feedback?}
\author{Akansha Kalra \\
        Kahlert School of Computing\\
        University of Utah \\
        akanshak@cs.utah.edu
        \And 
        Daniel S. Brown \\
        Kahlert School of Computing\\
        University of Utah \\
        daniel.s.brown@utah.edu }

\maketitle

\begin{abstract}
Reinforcement Learning from Human Feedback (RLHF) has emerged as a popular paradigm for capturing human intent to alleviate the challenges of hand-crafting the reward values. 
Despite the increasing interest in RLHF, most works learn  black box reward functions that while expressive are difficult to interpret and often require running the whole costly process of RL before we can even decipher if these frameworks are actually aligned with human preferences.
We propose and evaluate a novel approach for learning expressive and interpretable reward functions from preferences using Differentiable Decision Trees (DDTs).
Our experiments across several domains, including CartPole, Visual Gridworld environments and Atari games,  provide evidence that the tree structure of our learned reward function is useful in determining the extent to which the reward function is aligned with human preferences.
We also provide experimental evidence that not only shows that reward DDTs can often achieve competitive RL performance when compared with larger capacity deep neural network reward functions but also demonstrates the diagnostic utility of our framework in checking alignment of learned reward functions.
We also observe that the choice  between soft and hard (argmax) output of reward DDT reveals a tension between wanting highly shaped rewards to ensure good RL performance, while also wanting simpler, more interpretable rewards. Videos and code, are available at: \url{https://sites.google.com/view/ddt-rlhf}

\end{abstract}

\section{Introduction}
\begin{wrapfigure}[13]{r}{0.45\textwidth}
\vspace{-14mm}
  \begin{center}
\includegraphics[width=0.45\textwidth]{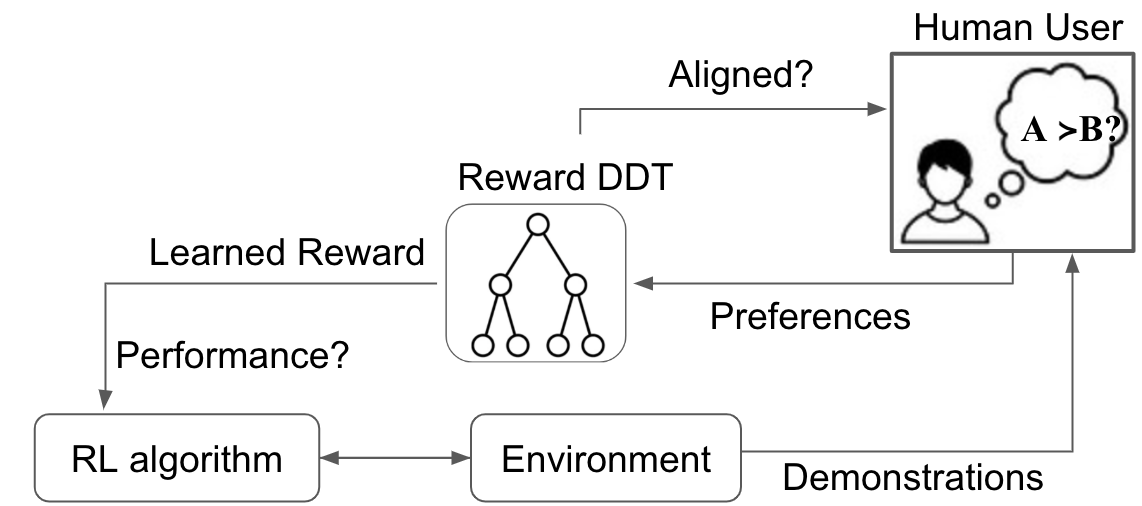}
    \end{center}
  \caption{We propose an end-to-end differentiable approach for training reward functions using differentiable decision trees via trajectory preference labels to enable interpretability and identification of misalignment of the learned reward function.}
  \label{fig:system_diag}
  \vspace{-10mm}
\end{wrapfigure} 

The reward function is central to reinforcement learning (RL) algorithms~\citep{sutton2018reinforcement}; however, it is difficult to manually specify a good reward function for many tasks~\citep{ng1999policy,krakovna2020specification}, motivating learning reward functions from human input~\citep{jeon2020reward}. We focus on the problem of learning interpretable reward functions. 

Most modern reward learning methods use deep neural networks ~\citep{finn2016guided,christiano2017deep,ibarz2018reward,henja2022fewshot,tien2023study}. However, despite the growing interest in explaining black box models trained via deep learning~\citep{gilpin2018explaining, zhang2018visual, heuillet2021explainability,raukur2022toward}, deep neural networks remain extremely difficult to interpret.  In the context of reward learning, it is especially critical that we can interpret the learned objective---if we cannot understand the objective that a robot or AI system has learned, then it is difficult to know if the AI system’s behavior will be aligned with human preferences and intent~\citep{russell2015research,leike2018scalable,brown2021value}. This is particularly significant in tasks where human safety is on the line, for example in healthcare, autonomous navigation, and assistive robots.

Thus, we are faced with a problem: we want highly accurate and expressive reward models, but we also want to be able to interpret the learned reward function. In particular, we seek to integrate structural and interpretability constraints into the reinforcement learning from human feedback (RLHF) pipeline to improve diagnostic capabilities for misalignment issues. A natural step towards both of these goals is to combine the expressiveness of neural networks with an architecture choice that is easier for a human to interpret, such as a decision tree. 
To tackle the the aforementioned problems, we propose a novel reward learning approach that uses an end-to-end differentiable decision tree model for learning interpretable reward functions from  pairwise preferences. 
We evaluate our approach on three different domains: CartPole~\citep{brockman2016openai}, a novel set of Visual MNIST Gridworld environments, and two Atari games from the Arcade Learning Environment~\citep{bellemare2013arcade}. We investigate the ability to learn expressive and interpretable reward functions from both low- and high-dimensional state inputs.  
 
Learning a reward model as a differentiable decision tree has the advantage that the tree structure explicitly breaks the reward prediction for a state into a finite number of routing decisions within the tree. This provides the potential to understand how the reward predictions are being made. Leveraging the tree structure, we can provide global explanations across both low- and medium-dimensional environments such as CartPole and visual MNIST gridworlds. For high-dimensional visual state spaces, such as Atari, we propose a novel form of hybrid explanation that seeks to provide global explanations by leveraging aggregations of individual input states. 

Our paper makes the following contributions:
(1) We introduce a reward learning framework (Fig ~\ref{fig:system_diag}) that employs differentiable decision trees (DDTs) to learn human intent using trajectory preference labels without necessitating any hand-crafting of the input feature space. \textit{To the best of our knowledge, our framework is the first interpretable tree-based method for reward learning that can be applied in visual domains.} (2) We propose hybrid explanations for internal nodes that approximate  global explanations by leveraging aggregations of individual input states. (3) We study the ability of DDTs to learn interpretable rewards on visual-control tasks and find that Reward DDTs can often learn interpretable reward functions. We also provide evidence that reward DDTs can be used to identify reward misalignment. (4) We find that the policies obtained by optimizing our reward DDTs via RL often perform comparably to policies trained with black-box neural network reward functions.

\section{Related Work}
\paragraph{Preference Learning} 

Reinforcement learning from human feedback (RLHF), is a common approach for learning reward functions and corresponding RL policies~\citep{wirth2016model}. 
It has been shown that preference learning allows generalizing to various domains, even when sub-optimal demonstrations are provided without any explicit preferences  and can achieve better-than-demonstrator performance~\citep{brown2019extrapolating}. 
Preference learning is also applicable across multiple forms of human input: prior work has shown that demonstrations~\citep{brown2020better}, e-stops~\citep{ghosal2022effect}, rankings~\citep{ouyang2022training}, and corrections~\citep{mehta2022unified}, can all be represented in terms of pairwise preferences. Thus, our approach is also applicable in these other settings. Prior work on RLHF typically either assumes access to a set of hand-designed reward features~\citep{sadigh2017active,erdem2020asking,mehta2022unified,ghosal2022effect} or uses deep convolutional or fully connected networks for reward learning~\citep{christiano2017deep,brown2019extrapolating,lee2021pebble,henja2022fewshot,ouyang2022training,liu2023efficient,karimi2024reward}. By contrast, we study the extent to which we can learn expressive, but also interpretable reward functions via differentiable decision trees~\citep{frosst2017distilling}.

\paragraph{Explaining and Interpreting Reward Functions}
In the  past few years, various attempts have been made to understand learned reward functions. Prior work compares learned reward functions to a ground truth reward using pseudometrics~\citep{gleave2020quantifying}, saliency maps and counterfactuals~\citep{brown2019extrapolating,michaud2020understanding,mahmud2023reveale,tien2023study}. Other work leverages human teaching strategies~\citep{lee2021machine,booth2022revisiting} or uses human-centric evaluation methods for reward explanation~\citep{9822391}. Prior work has also looked at using expert-driven reward design techniques to incorporate structural and interpretability constraints~\citep{jiang2021temporal,devidze2021explicable,icarte2022reward}. 
We seek to investigate to what extent differentiable decision trees enable interpretable reward functions.

\paragraph{Differentiable Decision Trees} 
Differential decision trees (DDTs) seek to combine the flexibility of neural networks with the logical and interpretable structure of decision trees~\citep{quinlan1986induction,jordanDT}. DDTs have been previously applied to supervised  learning tasks~\citep{frosst2017distilling, tanno2019adaptive, hazimeh2020tree} and unsupervised tasks ~\citep{zantedeschi2021learning}. Recent work has also investigated using DDTs for reinforcement learning tasks~\citep{silva2020optimization,coppens2019distilling,10105980,ding2021cdt,pace2022poetree}, but focuses on \textit{policy learning} using DDTs. Compared to prior work, the primary objective of our work is to \textit{learn interpretable reward functions} using DDTs.  While policy explanations are important, they only show what triggers an
agent to take a certain action, rather than explaining the underlying reason why the policy has learned to take
take an action. By understanding agent’s
reward function, we gain insight into the agent’s value alignment~\citep{leike2018scalable,fisac2020pragmatic,brown2021value}. Importantly, understanding an agent's reward function can enable an understanding of how that agent would act across different embodiments and dynamics~\citet{fu2018learning,zakka2022xirl}, unlike policies which are tied to the specifics of the MDP transition dynamics and action space. Furthermore, prior work using DDTs for policy learning only considers low-dimensional, non-visual inputs~\citep{silva2020optimization,coppens2019distilling}. By contrast, we study DDTs applied to high-dimensional image observations.

\paragraph{Decision Trees for Reward Learning}
There has been very little prior work on using decision trees for reward learning. \citet{bewley2022interpretable} recently pioneered the idea of a tree-based reward function. However, their approach to learning a tree-based reward requires a complex, non-differentiable, multi-stage optimization procedure. By contrast, our approach is end-to-end differentiable and trainable using a simple cross entropy loss. \citet{bewley2022interpretable} also only consider low-dimensional inputs where internal nodes in tree have the form $(s,a)_d \geq c$ for each dimension $d$ of the state-action space and threshold $c$. This approach divides  state-action space into axis aligned hyperrectangles, which often works for lower-dimensional spaces, but does not scale to higher-dimensional state and action spaces. Follow-on work~\citep{bewley2023reward} uses a differentiable loss function but is not end-to-end differentiable as it requires reward tree to regrow at each update and requires hand crafting input features per decision node in the tree, making it intractable to scale to the types of visual inputs we consider. 
We seek to extend the state-of-the-art in interpretable tree-based reward learning by learning reward function DDTs that are end-to-end differentiable, do not require hand-crafted features, and scale easily to high dimensional pixel inputs. 
 
\section{Reward Learning using Differentiable Decision Trees}
Classical decision trees are often interpretable and easy to tune~\citep{kotsiantis2013decision,molnar2020interpretable}; however, they require feature engineering which can result in lower performance and less generalization compared with other machine learning approaches~\citep{frosst2017distilling,hazimeh2020tree}. In this section, we discuss our proposed approach for learning interpretable but expressive reward functions via differentiable decision trees (DDTs). 

While classical decision trees consist of internal nodes that deterministically route inputs, we want our reward function tree to be easily trained using backpropagation. Thus, we need a differentiable soft routing function that retains the expressiveness of a neural network by learning the routing function for each non-leaf node.
We define an internal node in the DDT as a sequence of one or more parameterized functions applied, to the input to the DDT to determine probability of routing left or right. To facilitate interpretability, each internal node depends directly on the input---this is a common design choice in DDTs~\citep{frosst2017distilling} and serves our purpose well by allowing us to easily trace each routing decision in the tree to the raw input features. Thus, the differentiable decision tree learns a hierarchy of decision boundaries that determine the routing probabilities for each input. We describe two variants of an internal node below:

\subsection{Internal Nodes}
\begin{wrapfigure}[7]{r}{0.29\textwidth}
\vspace{-15mm}
  \begin{center}
  \includegraphics[width=0.3\textwidth]{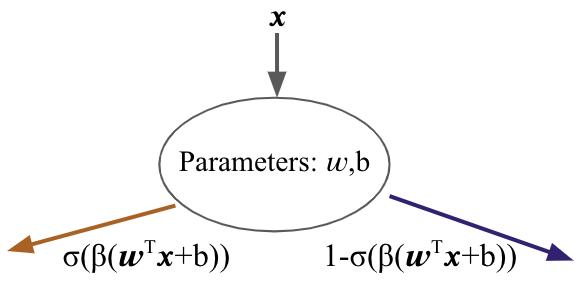}
    \end{center}
  \caption{Routing probability of an internal node in a DDT.}
  \label{fig:internal_node}
\end{wrapfigure}

 \paragraph{Simple Internal Node}
 Proposed by \citet{frosst2017distilling}, a simple internal routing node, ${i}$, has a linear layer  with learnable parameters $\textbf{w}_{i}$ and a bias term $b$ upon which a sigmoid activation function, $\sigma$, is applied to derive the routing probability given an input \textbf{x} (Fig~\ref{fig:internal_node}). Thus, the probability at node $i$ of routing to the left branch  is defined as 
    $p_{i}(\textbf{x})= \sigma(\beta(\textbf{x}\cdot \textbf{w}_{i}+b))$.
The inverse temperature parameter, $\beta$, controls the degree of soft decisions.

\paragraph{Sophisticated Internal Node} 
For higher-dimensional inputs we propose  an alternative  internal node architecture, which consists of a single convolutional layer with Leaky ReLU as the non-linearity followed by a fully connected linear layer, as before. The probability of going to the leftmost branch at an internal node $i$ is defined as 
    $p_{i}(\textbf{x})= \sigma({
   (\text{LeakyReLU}
    (\text{Conv2d}(\textbf x)))} \cdot
    \textbf{w}_{i}+b)$.

\subsection{Leaf Nodes}
 
 Following prior work that uses DDTs for classification problems~\citep{frosst2017distilling}, we parameterize each leaf node,  $l$, with a learnable parameter vector $\bm{\phi}^{l}$, that defines a softmax distribution over a discrete number of classes $c$. The probability distribution,  $\bm{Q}^{l}$, over outputs at a leaf is defined as $\bm{Q}^{l}_{i} = \exp(\bm{\phi}^{l}_{i})/(\sum_{j=0}^c\exp (\bm{\phi}^{l}_{j}))$. We propose two ways to obtain rewards at the leaf nodes:

 

\paragraph{Multi-Class Reward Leaf (CRL)} This formulation of leaf node performs \textit{multi-class classification} and assumes that the user specifies a set of $c$ unique  discrete reward values that the DDT can output in the form of a vector
$\textbf{R} = (r_1, r_2, \ldots, r_{c})$, where $c$ denotes the number of classes for the DDT, and each class index $i$ is assigned reward value $r_i$. Thus, the learnable parameters, $\bm{\phi}^{l}$, at multi-class reward leaf $l$ form the logit values of a classification problem over the possible reward values in $\textbf{R}$.

\paragraph{Min-Max Reward Interpolation Leaf (IL)} As an alternative to the classification approach, we also propose to model  the reward  of a DDT as  \textit{regression problem}, that only requires the user to specify the minimum and maximum range of possible reward values as opposed to requiring finite set of  possible reward values as in CRL. Thus, $c=2$ and the reward vector is of the form $\textbf{R} = (R_{\min}, R_{\max})$, where $R_{\min}$ and $R_{\max}$ correspond to minimum and maximum desired reward output, respectively. Given this parameterization, we interpret the reward output of a DDT leaf node as a convex combination of $R_{\min}$ and $R_{\max}$ based on the learned parameters $\bm{\phi}^l$. 



\subsection{ Training DDTs for Reward Learning using Human Preferences}

As we want our reward DDT to be end-to-end differentiable when learning a reward function from preference labels, we need to find a way to formulate 
soft reward prediction. Given a tree of depth $d\geq1$, we have $\sum_{k=0}^{d-1} 2^{k}$ internal nodes and $2^d$ leaves. To formulate a differentiable objective, we first denote the path probability, given an input \textbf{x}, from the root node to a leaf $\ell$ by $P^{\ell}(\textbf{x})$. The soft reward prediction of the tree is given by the sum over all leaves, $\ell$, of the path probability of reaching each leaf, $P^{\ell}(\textbf{x)}$,  multiplied with the soft reward output at that leaf:
\begin{equation}
    r_\theta(\textbf{x})=\sum_{\ell}P^{\ell}(\textbf{x}) (\textbf{Q}^{\ell} \cdot \textbf{R}) \;.
\end{equation}
To train our reward function DDT, we propose to leverage pairwise preference labels over trajectories. Given preferences over trajectories of the form $\tau_i \prec \tau_j$, where $\tau=(\textbf{x}_{1},\textbf{x}_{2},...\textbf{x}_T)$, we can train our entire differentiable decision tree via the following cross entropy loss  resulting from the Bradley Terry model of preferences~\citep{bradley1952rank,christiano2017deep,brown2019extrapolating}:
\begin{equation}
 \mathcal{L}(\theta) = -\sum_{\tau_i \prec \tau_j} \log \frac{\exp \displaystyle\sum_{\textbf{x} \in \tau_j} r_\theta(\textbf{x})}{\exp \displaystyle\sum_{\textbf{x} \in \tau_i} r_\theta(\textbf{x}) + \exp \displaystyle\sum_{\textbf{x} \in \tau_j} r_\theta(\textbf{x})} \; .
 \label{eqn:BTloss}
\end{equation}


\subsection{Using a Trained Reward DDT for Reward Prediction}
Given a trained reward DDT, we want to optimize the learned reward using RL.
One option is to use the soft reward (averaged across all leaf nodes weighted by routing probability); however, this loses interpretability since we cannot trace the predicted reward to a small number of discrete decisions. To enable interpretable reward predictions, we can alternatively output a single reward prediction by first finding the leaf node with maximum routing probability for a given input \textbf{x}:
\begin{equation}
    l^* = \arg\max_{\ell \in L}  P^{\ell}(\textbf{x}) \;, 
\end{equation}
where $L$ denotes set of all leaf nodes in the DDT. The test-time output of a reward DDT with a multi-class reward leaf (CRL) nodes is given as 
     $r_{max}(\textbf{x})= r_i, \text{ for }\; i = \arg\max_{i}\textbf{Q}^{\ell^*}_i;$
while for a reward DDT with min-max interpolation leaf (IL) nodes the reward output is given as
     $r_{max}(\textbf{x})= \textbf{Q}^{\ell^*} \cdot (R_{\min}, R_{\max})$.



\subsection{ Hybrid Explanations of Learned Reward DDT} \label{subsec:traces}
Depending on the dimensionality of the state space in a given environment, our framework allows us to create global explanations across all inputs in form of node activation heatmaps (discussed in further detail later). As an alternative, we also investigate hybrid explanations that approximate global explanations by leveraging aggregations of input states to visually understand the routing probability of each internal node. Inspired by \cite{bobu2022inducing}, we do this by visualizing a synthetic trace at each internal node. The synthetic trace is a sequence of states sorted by the probability of being routing left in decreasing order---the trace begins with the state that has maximum probability of being routed left and ends with the state that has minimum probability of being routed left.

\section{Experiments and Results}
We designed our experiments to investigate the following questions: (1) Can we detect misalignment in reward function by learning the reward function as a DDT? (2) How does modeling a reward function as a DDT influence downstream RL performance? (3) How does the choice of leaf node (multi-class reward leaf (CRL) or min-max reward interpolation leaf (IL)) affect performance?  (4) How does an increase in the environment complexity impact our design choices as well as our ability to interpret the learned reward function? 
To explore and address these questions, we perform evaluations on three different types of environments: CartPole, a novel set of MNIST Gridworld environments, and Atari 2600 games~\citep{ale}. 

We use CartPole to perform an initial assessment of our framework and provide an example of how interpreting a learned reward DDT enables detection of a \textit{silent misalignment problem}---the reward function is misaligned but the policy still performs well.
To evaluate our framework's ability on visual domains, we explore two MNIST Gridworld environments of increasing complexity, where the gridworlds have image based observations. Finally we examine our framework on Atari where the true score is masked and the agent must learn a reward function by interpreting high-dimensional pixel observations derived from video frames. The Atari domains provide evidence in higher-dimensional environments of the ability to detect reward misalignment.



\subsection{CartPole} 

\begin{figure}[t]
     \centering
         \vspace{-2mm}
         \includegraphics[width=0.88\linewidth]{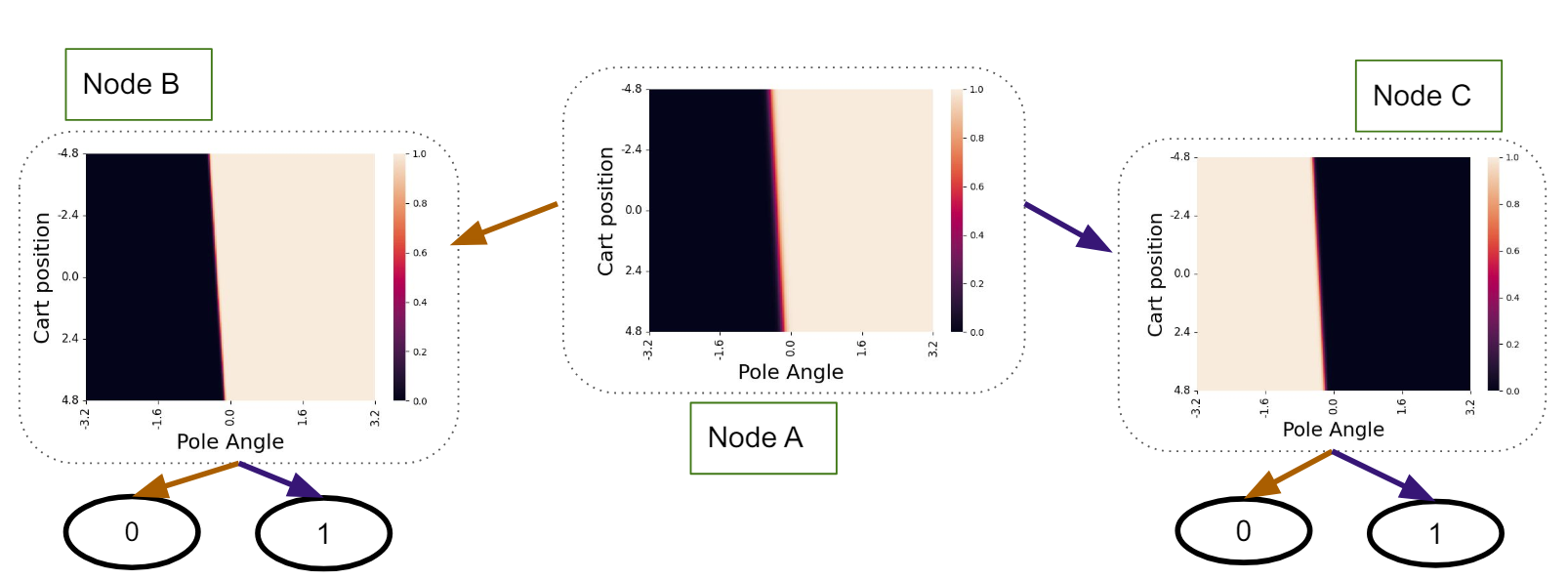}
         \caption{\textbf{Identifying Misalignment in the CartPole Reward DDT.} The heatmap for each internal node depicts the learned routing probability. Leaf nodes are depicted as circular nodes with their soft reward values. The tree learns that small magnitude pole angles are good and should be routed to a +1 reward but there is no learned decision boundary that clearly captures the preference that cart position stay within the range $[-2.4,2.4]$ showing that learned reward is misaligned due to the bias in the training dataset---the cartpole usually falls over long before the cart runs off the track. }
        \label{fig:Visualizing CartPole Interpretability}
\end{figure}

The CartPole environment comprises a cart with a pole attached to it, sliding on a friction-less track~\citep{brockman2016openai}. For this task, we wish to teach the agent to balance the pole on the cart for as long as possible while cart moves to left and right along the track without letting pole fall beyond $\pm 12^{\circ}$ from the upright position and without letting the cart move beyond $\pm 2.4$ units along the track. We assume no access to the true reward and must learn this from trajectory preferences.

\vspace{-3mm}
\paragraph{Setup} To train a reward function DDT, we generate trajectories by running a random policy in the environment for 200 steps for each trajectory. Following the advice of \citet{freire2020derail}, we remove the standard terminal or done flag to avoid leaking information about the true reward. The terminal flag normally is triggered in CartPole when either the pole falls or the cart goes off the track. Instead, we make CartPole a fixed horizon task by always accumulating states in each trajectory for 200 timesteps---even if the pole falls over. 
We design a synthetic preference labeler that returns pairwise preferences based on the true (but unobserved) reward of +1 only if the cart position $x \in[-2.4,2.4]$ and the pole angle $\theta \in [-12^{\circ},+12^{\circ}]$ and $0$ otherwise. Pairwise preferences are assigned based on the true reward for each trajectory.

Given pairwise preference labels over suboptimal trajectories, we train a reward DDT with 3 internal nodes and 4 leaf nodes. 
We use multi-class reward leaf (CRL) nodes with 2 classes: $\textbf{R} = (0.0, 1.0)$
 (for more details, refer to Appendix~\ref{app:cartpole}).
It is important to note that even though the ground truth preferences are based on both cart position and pole angle, the pole usually falls past the desirable range long before the cart leaves the desirable range. Thus, our dataset is biased and may lead to a misaligned reward function.
We evaluate RL performance of the learned reward DDT, by running PPO on the learned reward function to obtain the final policy and then evaluate this learned policy on the ground-truth reward function. We also compute the performance of a PPO policy trained on  the same dataset using a neural network reward function.
To unveil the fact that our learned reward functions (using both DDT and neural network) are biased, we run RL experiments in two settings: (1) \textit{In-Distribution} uses the default starting cart position in the range $[-0.05, 0.05]$ as in our training dataset and (2) \textit{Out-Of-Distribution} where the starting cart position is in the range $[2.35,2.45]$ (the boundary of the range of desired track positions).

\begin{table}
\footnotesize
  \caption{\textbf{Silent Misalignment in CartPole.} CRL denotes Class Reward Leaf nodes. 
  For In-Distribution, DDTs with soft outputs and argmax rewards perform on par with a non-interpretable fully connected 2-layer reward network baseline and with RL policy learned under ground truth reward. For Out-Of-Distribution, the RL policy of  learned reward models, both DDT and neural network fails to learn to balance pole while moving along the track while RL policy under ground truth reward learns to balance pole as it moves on track. 
  The table shows Mean and Standard deviation across 10 seeds averaged over 100 rollouts as well as the Interquartile Mean (IQM).}
  \label{CP-table}
  \centering
    \begin{tabular}{llcccc}
    \toprule
    &&\multicolumn{2}{c}{DDT} & \multicolumn{2}{c}{Baselines} \\
    && CRL Soft    & CRL Argmax  & Neural Network & Ground Truth\\
    \midrule

    \multirow{2}{*}{In-Distribution} & Mean (Std) & 190.9 (28.1)  & 200.0 (0.0)  & 156.3 (59.0)   & 200.0 (0.0)  \\
    &IQM  & 200.0  & 200.0 &   179.5 & 200.0  \\   
    \hline
    \multirow{2}{*}{Out-Of-Distribution} & Mean (Std) & 8.8 (3.7)  & 7.7 (2.1) & 20.7 (39.2)    & 172.0 (45.6)  \\
    &IQM  & 8.3  & 7.9 &   8.8 & 185.3  \\
    \bottomrule
  \end{tabular}
\end{table}


\paragraph{Results}
The In-Distribution results in Table~\ref{CP-table} show that RL performance of a simple reward DDT is comparable to that of a neural network made up of fully-connected layers as well as to RL policy learned under ground truth reward, irrespective of whether the policy is learned using soft rewards or using the maximum probability path across the learned reward DDT. This primarily gives us the evidence that our framework can achieve relatively competitive performance as that of a neural network for state based observations,before we move on to image-based observations.

Fig~\ref{fig:Visualizing CartPole Interpretability} shows learned reward DDT. Because the input space to the reward function is 2-dimensional (cart position and pole angle) we visualize the heatmap of  routing probability at each internal node (as a function of cart position and pole angle) along with leaf distributions. From DDT it is clear that most of the routing decisions are made based on pole angle, rather than cart position. 
A nice feature of the reward DDT is that we can easily visually interpret the learned reward just by looking at the tree. From Fig~\ref{fig:Visualizing CartPole Interpretability} we see that while the tree learns that small magnitude pole angles are good and should be routed to a +1 reward, there is no learned decision boundary that clearly captures the preference that cart position stay within the range $[-2.4,2.4]$. We call this a \textit{silent misalignment problem}. Similar to a silent bug in programming, it is not obvious by running RL that anything is wrong with the learned reward function---it turns out that trying not to tip the pole is a decent surrogate reward function that works well in the standard CartPole environment. Thus, the agent has learned the right policy for the wrong reason, something that is only clear by interpreting the learned reward. While this poses no serious issues in the standard CartPole environment, silent alignment problems could lead to unwanted behavior under distribution shifts and detecting these silent alignment problems is an open challenge in AI safety and alignment research~\citep{ji2023ai}.

Indeed, the Out-Of-Distribution results in Table~\ref{CP-table} demonstrate this silent misalignment in the learned reward functions, where the policies learned from the reward DDTs as well as neural network reward learn to balance the pole, but fail to stay in the desired track range. In contrast, the RL policy under ground truth reward learns to balance pole correctly while moving along the track, starting from any state. 
Our DDT  framework makes it easier to detect this misalignment in learned reward function prior to running RL, but with non-interpretable black box neural network's learned reward function we had to incur cost of running RL before we could uncover the bias in the learned reward.

\subsection{MNIST Gridworlds}

\begin{wrapfigure}[8]{r}{0.25\textwidth}
\vspace{-13mm}
  \begin{center}
\includegraphics[width=0.25\textwidth]{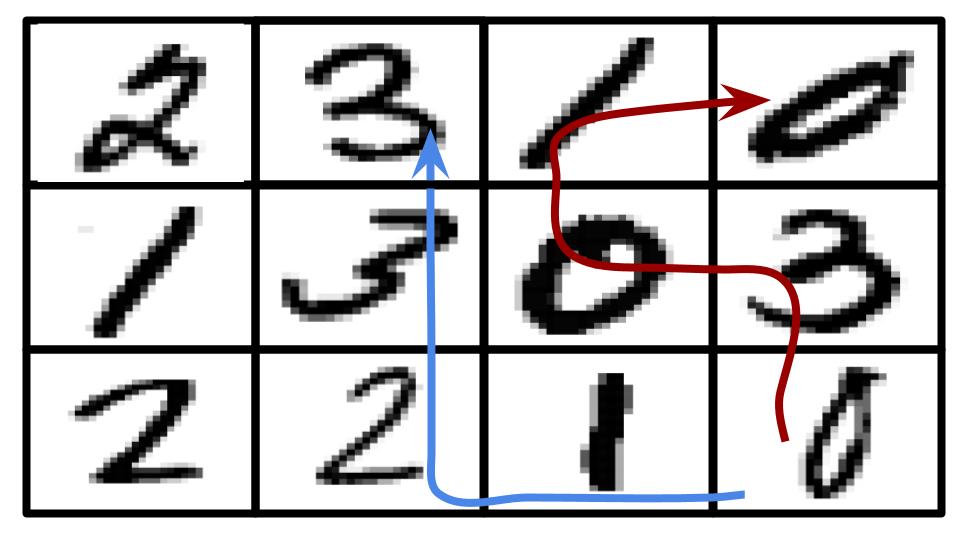}
   \end{center}
   \vspace{-5mm}
  \caption{MNIST Gridworld with a pair of trajectories where the blue trajectory is preferred.}
  \label{fig:grid03}
\end{wrapfigure}

Next we evaluated our reward DDT framework on two novel MNIST gridworld environments of increasing difficulty. In each environment the agent can move in the 4 cardinal directions and each state is associated with a $28 \times 28$ grey-scale image of the MNIST digit and the value of the digit determines the true unobserved reward at that state (for more details, refer to Appendix~\ref{app:mnist_val}) .

The true reward is unobserved and must be inferred from preferences over pairwise preferences over trajectories.
To interpret the learned reward DDT, we construct a pixel-level activation heatmap for each internal node by starting with a blank image and iteratively toggling on and off each pixel and computing the resulting difference in routing probabilities for each internal node. We compare the performance of a policy optimized using the learned DDT reward function against the optimal policy under the true reward, a random policy, and a policy learned by optimizing a black-box neural network reward function trained on the same preference dataset. We also report accuracy of the learned reward models on validation set of pairwise preferences over trajectories in Appendix~\ref{app:mnist_val}.

\vspace{-2mm}
\subsubsection{MNIST (0-3) Gridworld}\label{sec:mnist0-3}

\paragraph{Setup} We begin by examining our framework for image based inputs on a simple 5x5 gridworld where each state in the MDP corresponds to a MNIST digit 0, 1, 2, or 3 (see Fig~\ref{fig:grid03} for an example pairwise trajectory comparison).
We trained reward DDTs of depth 2 with 3  simple internal nodes and  4  leaf nodes either all of type CRL with $\textbf{R} =(0,1,2,3)$ and or all of type IL with $R_{\min}= 0$ and $R_{\max}=3$ using a learning rate of $0.001$ and weight decay $0.005$ and the Adam optimizer. 
\begin{figure*}[t!]
\centering
  \begin{subfigure}[]{0.44\linewidth}
    \centering
    \includegraphics[width=\textwidth]{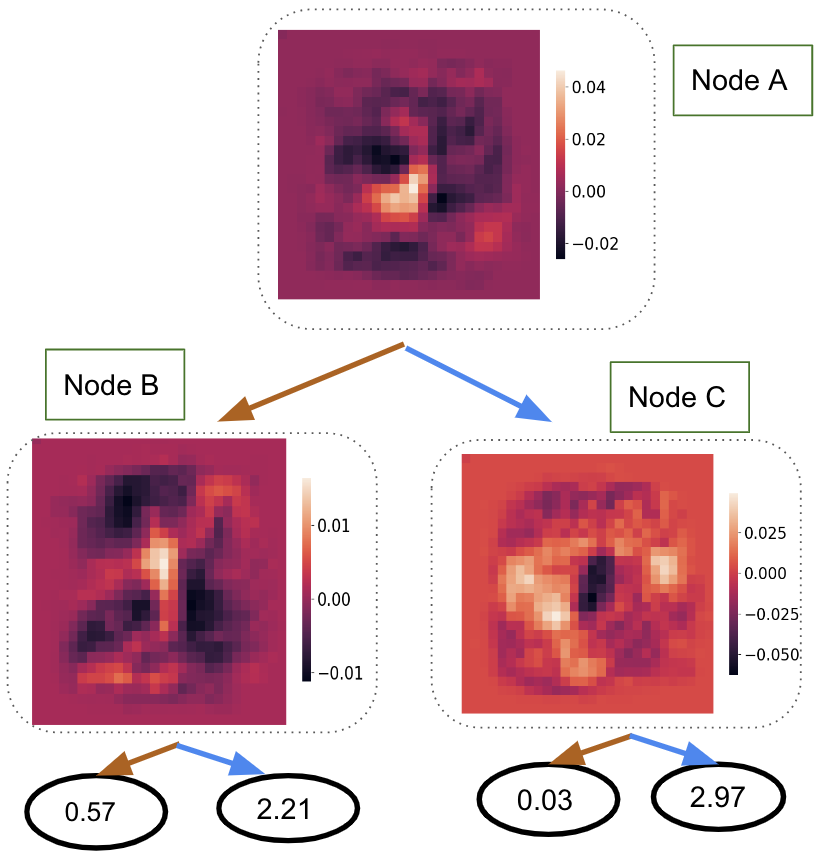}
   \caption{}
        \label{fig: Long MNIST 0-3 Reward DDT}
   \end{subfigure}
   \hfill
     \begin{subfigure}[]{0.44\linewidth}
         \centering
         \includegraphics[width=\textwidth]{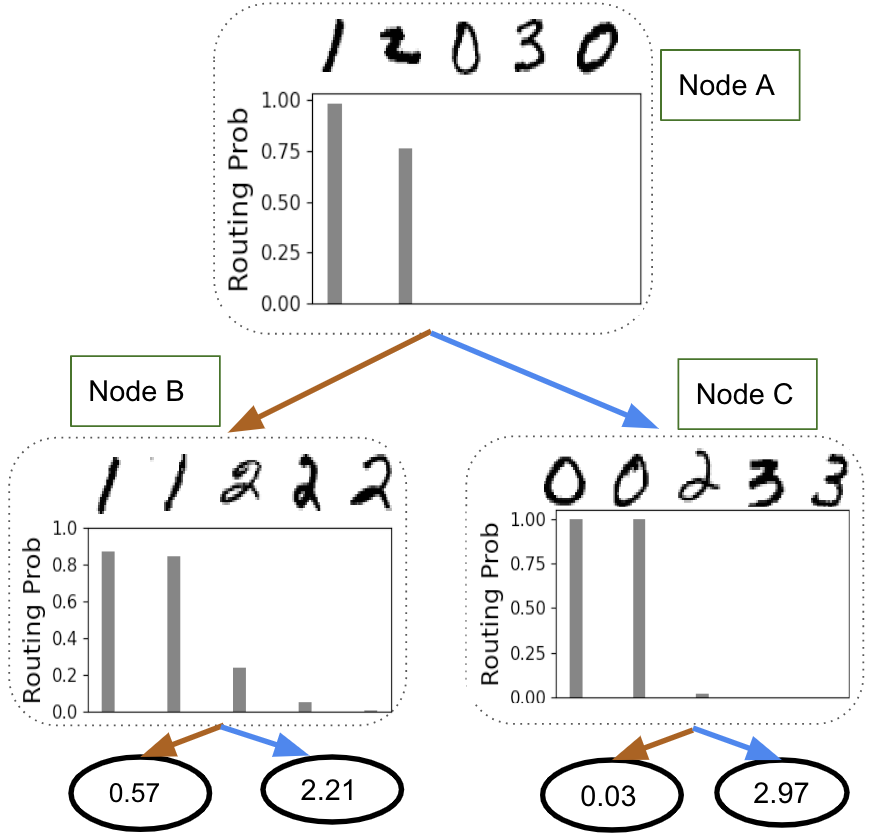}
        \caption{}
    \label{fig:0-3traces}
     \end{subfigure}

    \caption{\textbf{Interpreting the MNIST (0-3) Interpolated Leaf Reward DDT}. Leaf nodes are depicted as circular nodes with their learned reward values. \textbf{(a)} Visualization of activation maps give insight into the learned routing of the DDT.  \textbf{(b)} Visualization of synthetic traces along with their respective routing probabilities. }
    \label{fig:MNIST0-3}
\end{figure*}

\begin{table*}[t!]
\footnotesize
  \caption{RL Performance as the
percentage of expected return obtained relative to the performance of an optimal policy on the ground-truth reward. Results are averaged across 100 different MDPs. We find evidence that reward DDTs with Interpolated Leaf nodes (IL)  perform similar to neural network reward functions, while using Class Reward Leaf nodes (CRL) results in much lower performance, but still outperforms a random policy (Random). These results provides evidence that DDTs can learn both interpretable reward functions without causing a large degradation in RL performance. }
  \label{MNIST-table}
  \centering
  \begin{tabular}{lrrrrrr}
    \toprule
    &\multicolumn{4}{c}{Reward DDT} & \multicolumn{2}{c}{Baselines}                 \\
    & CRL Soft    & CRL Argmax   & IL Soft  & IL Argmax  & NNet & Random\\
    \midrule
    MNIST 0-3  & 71.7\% & 71.7\% & 98.06\% & 95.27\% & 99.5\% & 7.6\% \\
    MNIST 0-9 & 79.6\% & 79.6\% & 97.3\% & 92.9\% & 97.7\% & 7.9\% \\
    \bottomrule
  \end{tabular} 
  
\end{table*}


\paragraph{Results}
We visualized and compared the reward DDTs with CRL and IL leaf nodes and found that in CRL formulation the leaf nodes fail to specialize and the argmax output of the leaf nodes is either 0 or 3, despite investigating several regularization techniques (see Appendix~\ref{app:mnist_grid} for details and visualizations).
This provides evidence that using IL leaf nodes is better when learning complicated reward functions where we wish to output more than two possible rewards. Table~\ref{MNIST-table} also provides empirical evidence supporting the user of IL nodes. IL nodes are also simpler, as they only require specifying a range of desired reward output values, $[R_{\min}, R_{\max}]$. Thus, we focus on our analysis on the interpretability of the IL reward DDT.


Interestingly, we see in Fig~\ref{fig: Long MNIST 0-3 Reward DDT} that the activation heatmaps 
isolate pixel features that are maximally discriminative and aid in understanding of what the reward DDT has learned. 
These heatmaps show that DDT learns to route based on visual representations of each digit:  
Node B has learned to discriminate between 1's and 2's by assigning a high routing probability left for vertical pixels in the center (corresponding to the vertical stroke of the digit 1), while using upper and lower curves of digit 2 to route 2's right (note the black shadow that looks like a 2). 
Node C discriminates between digits 0 and 3 based on the middle cusp of 3 and left curve of the 0. 
Finally, node A learns to route 1's and 2's left and 0's and 3's right based on the presence of central lower pixels---the highest activation for node A is intersection of the 1 and 2 which falls between middle and lower cusps of 3 and inside digit 0. 
Despite the lack of fine-grained feedback and no explicit reward labels, when using min-max reward interpolation between $R_{min}=0$ and $R_{max}=3$, the DDT learns a close approximation to the actual state rewards and the learned rules in DDT are visually interpretable. 

We also interpret the same reward DDT using synthetic traces as shown in Fig~\ref{fig:0-3traces}. As described in Section~\ref{subsec:traces}, these traces approximate global explanation by leveraging aggregations of input states for each internal node. Each trace is a sequence of states sorted by the probability of being routed left in decreasing order. 
We find visual evidence that the DDT has learned to route digits with a vertical stroke to Node B which then discriminates between 1 and 2s, while digits with a circular curve form often get routed to Node C which then discriminates between digits 0, 2 and 3s. 



Row 1 of Table ~\ref{MNIST-table} shows that RL performance of using a reward DDT trained with IL leaf nodes exceeds the performance when using the classification-based CRL leaf nodes, both when running RL using soft reward outputs and when using the output of the maximum probability path in the tree (argmax). We also found that the learned reward DDT with CRL leaf nodes learns very high/low routing probabilities at each internal node and thus yields nearly identical reward values in both soft and argmax reward setting.
Moreover, RL performance of IL reward DDT using soft reward is only slightly lower than the performance of a deep neural network reward function. 
In Appendix~\ref{app:mnist_grid}, we compare the reward DDT in Fig~\ref{fig: Long MNIST 0-3 Reward DDT}, that is learned from pairwise preferences, with a DDT trained with explicit reward labels and a classification loss and find no significant degradation in interpretability from using pairwise preferences. 

\subsubsection{MNIST (0-9) Gridworld}
\paragraph{Setup} To assess the scalability of our framework, we next explored a 10x10 gridworld with state space comprising of MNIST digits 0 to 9. To further study the effects of leaf node type, we used reward DDTs of depth 4 with simple internal nodes and trained them with one of two types of leaf nodes: either CRL nodes  with $\textbf{R} = (0,1,..,9)$ or IL nodes with $R_{\min}= 0$ and $R_{\max}=9$.

\paragraph{Results} 
Row 2 of Table~\ref{MNIST-table} shows the IL soft reward performance is very similar to the performance of a black-box ConvNet learned reward. 
However, we find that performance of CRL softmax and argmax is significantly degraded, but much better than a random policy. This provides further evidence simply framing DDT learning as a classification problem is in sufficient for learning good reward function and that the flexibility of interpolation leaf nodes (IL) to learn real valued reward outputs helps with both interpretability and downstream RL performance.

This provides evidence that
our framework maintains high performance for much longer
horizon and more difficult tasks when using interpolated leaf nodes (IL). 
Even though tree structures can help with interpretability, the deeper the tree, the harder it is to understand what is going on. In Figure~\ref{fig:whole mnist traces} in the Appendix, we visualize the learned IL Reward DDT. While there are some noticeable trends, it is also hard to interpret exactly how the DDT has learned to route nodes. Thus, while our results provide evidence that high-performing policies can be learned via RLHF using reward DDTs, the more complex the DDT, the more difficult it is to interpret.



\subsection{Atari} \label{Atari-main}
As a final test of the efficacy and scalability of learning interpretable rewards via DDTs, we trained reward DDTs on the Beam Rider and Breakout Atari games~\citep{ale}. Learning rewards for these games is challenging as the states are high-dimensional pixel inputs consisting of stacks of four $84\times 84$ video frames and many prior works have used Atari games to study reward function learning~\citep{christiano2017deep,tucker2018inverse,ibarz2018reward,brown2019extrapolating}. 

\vspace{2mm}
\noindent\textbf{Setup} To train our reward DDT, we used the open-source offline preference datasets collected by~\citet{brown2019extrapolating}\footnote{\url{https://github.com/hiwonjoon/ICML2019-TREX}}.
We then examine whether a reward DDT can match the RL performance of T-REX, a deep convolutional neural network offline RLHF approach proposed by \citet{brown2019extrapolating}, while also being interpretable.
Because of the complexity of the task, we use sophisticated internal nodes and IL leaf nodes with $R_{\min} =0$ and $R_{\max}=1$ (see Appendix~\ref{app:atari} for full details).


\begin{figure*}
\centering
  \begin{subfigure}[t]{0.49\linewidth}
    \centering
    \includegraphics[width=\linewidth]{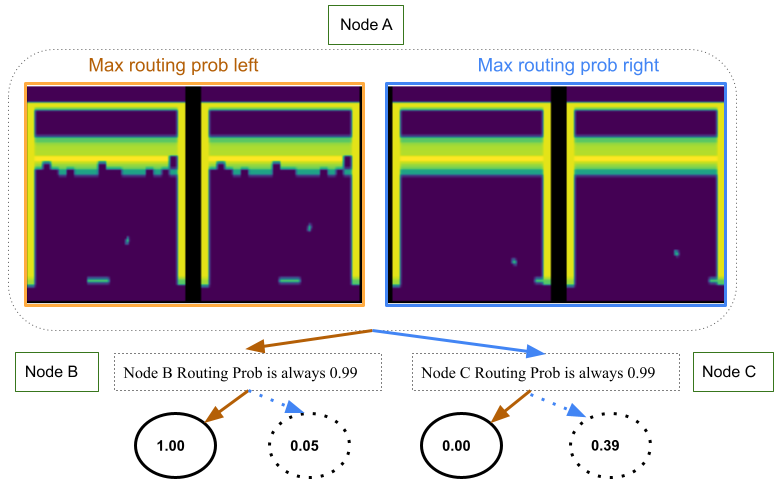}
    \caption{DDT without activation penalty regularization}
    \label{fig:Breakout_wo_penalty}
   \end{subfigure}
    \hfill
     \begin{subfigure}[t]{0.49\linewidth}
         \centering
         \includegraphics[width=\textwidth]{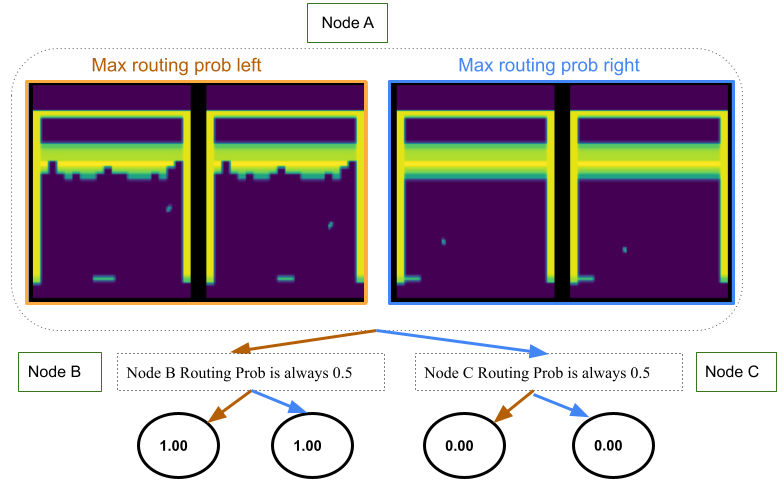}
         \caption{DDT with activation penalty regularization}
         \label{fig:Breakout_penalty}
     \end{subfigure}

    \caption{\textbf{Visualization of Breakout Reward DDTs}. We plot the DDTs trained without (a) vs with (b) a regularization penalty on the internal node routing probabilities. Dashed lines denote leaf nodes that are never reachable.}
    \label{fig:Breakout_without vs with penalty}
\end{figure*}


Because generated heatmaps for Atari, have been shown to have mixed results~\citep{brown2019extrapolating}, we opt to use traces for interpreting the learned reward DDTs. As before, the trace for an internal node begins with the state that has maximum probability of being routed left and ends with the state that has minimum probability of being routed left. For ease of visualization, we show the first and last state in the trace and find that they still provide useful information about the internals of the learned reward function.



%

\begin{figure*}
\centering
  \begin{subfigure}[t]{0.49\linewidth}
    \centering
    \includegraphics[width=\textwidth]{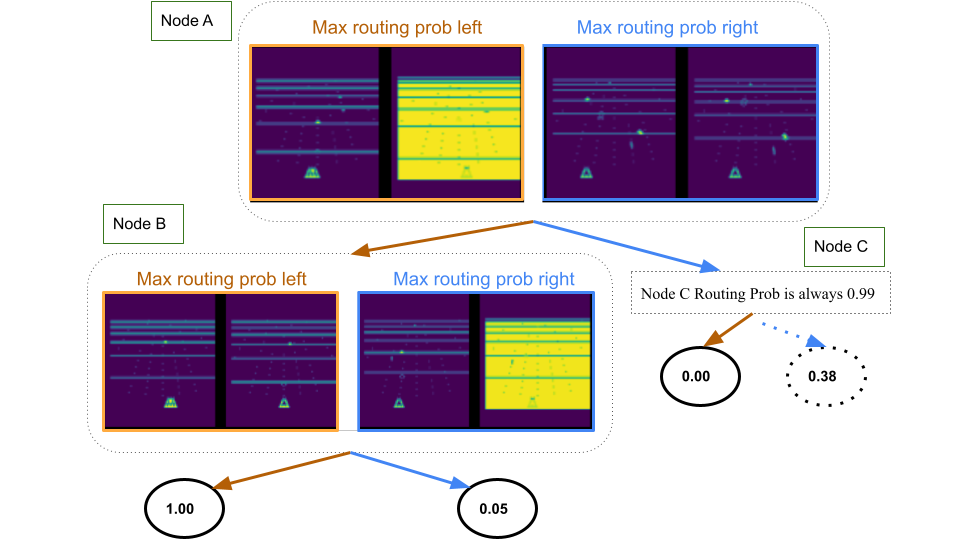}
    \caption{DDT without activation penalty regularization}
    \label{fig:BR_wo_penalty}
   \end{subfigure}
    \hfill
     \begin{subfigure}[t]{0.49\linewidth}
         \centering
         \includegraphics[width=\textwidth]{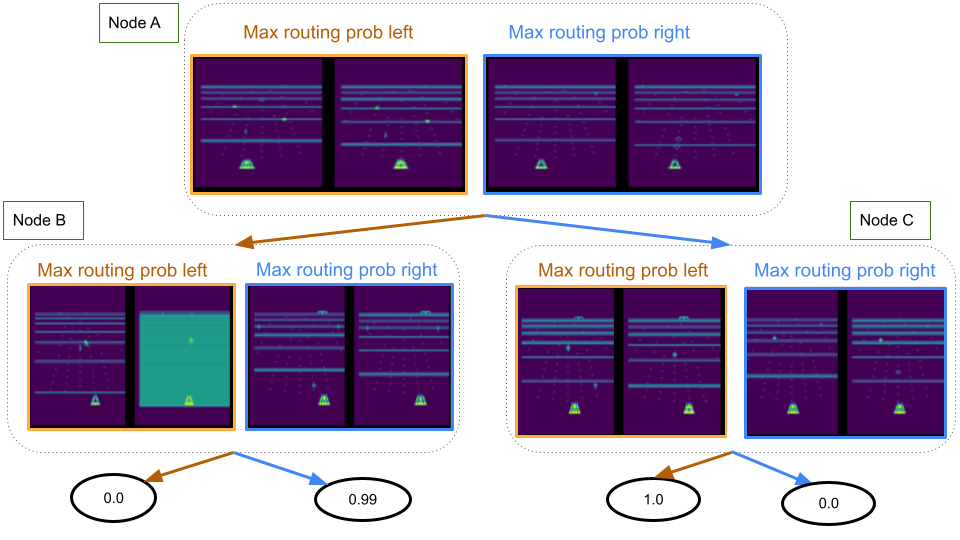}
         \caption{DDT with activation penalty regularization}
         \label{fig:BR_penalty}
     \end{subfigure}

    \caption{\textbf{Visualization of Beam Rider Reward DDTs}. We plot the DDTs trained without (a) vs with (b) a regularization penalty on the internal node routing probabilities.}
    \label{fig:BR_without vs with penalty}
\end{figure*}

For each learned reward, we optimized a policy by training an A2C~\citep{mnih2016asynchronous} agent using Stable Baselines 3 for 10 million timesteps.
As done in our previous experiments, when training the RL agent, we utilize each learned reward DDT in two ways: we either obtain a soft reward over all leaves from tree or we choose the path with maximum routing probability and the reward in this case is obtained by argmaxing over the maximum probability path.
We report mean and standard deviation across 10 seeds evaluated for 100 episodes each as well as the inter-quartile mean (IQM), which has been proposed as a better alternative when evaluating smaller numbers of seeds 
as recommended by prior works~\citep{patterson2023empirical,agarwal2021deep}.

\begin{table*}[t]
\footnotesize
  \centering
  \begin{tabular}{lrrrrr}
    \toprule
    &\multicolumn{4}{c}{DDT}   & Baseline                \\
     & $\neg$penalty&  $\neg$penalty & penalty & penalty& T-REX  \\
     Game& $\neg$argmax&  argmax & $\neg$argmax & argmax&   \\
    \midrule
    Breakout: Mean(Std) & 20.2 (34.4)   & 50.0 (105.3) & 83.5 (130.9) & 51.5 (100.6) & 58.3 (42.4) \\ 
    Breakout: IQM & 12.8 & 16.6 & 29.3 & 15.0 & 48.9 \\
    
    Beam Rider: Mean (Std) & 237.2 (322.2) & 189.4 (284.6) & 39.9 (40.7) & 107.6 (288.2) & 323.1 (335.9) \\ 
    Beam Rider: IQM & 94.9 & 64.1 & 30.2  & 7.6  & 254.9 \\
    
   \bottomrule
  \end{tabular}
  \caption{\textbf{Reinforcement learning using reward DDTs.} We report mean, standard deviation (Std) and inter-quartile mean (IQM) across 10 different seeds of RL evaluated for 100 epsiodes each.}
    \label{tab:rl-results}
\end{table*}

\vspace{2mm}
\noindent\textbf{Results}

While trying to create synthetic traces, we discovered that some of the leaf nodes were never reachable (e.g., the right child of Node B and C in case of Breakout Fig~\ref{fig:Breakout_wo_penalty} and the right child of Node C in case of BeamRider Fig~\ref{fig:BR_wo_penalty}).
We re-trained the sophisticated reward DDT with the same hyperparameters, but with an added penalty regularization to ensure that, on average across many inputs, each internal node routes left and right equally often across both environments (see Appendix~\ref{app:A_penalty} for details). 

We create a synthetic trace for the unregularized reward DDT for Breakout Fig~\ref{fig:Breakout_wo_penalty} by visualizing states that are routed with maximum and minimum probability to left and found evidence that states that have more bricks missing are routed left to Node B while the states in Node A that have few bricks missing have a lower routing probability and thus are routed right to Node C.
Both child nodes of the root node only use their respective left leaves and do not route any state to their respective right leaves, thus a synthetic trace could not be visualized for either Node B or Node C. Fig.~\ref{fig:Breakout_penalty} shows a similar trend, where the DDT learns to reward missing bricks. Interestingly, we did not find any evidence that the reward DDT learned to recognize the event of the ball hitting a brick. Instead of learning the causal ground truth reward that provides a reward each time a brick is hit, the reward DDT exhibits causal confusion~\citep{tien2023study} by learning to reward missing bricks. Similar to CartPole, we describe this as a ``silent misalignment problem". The reward function has learned to reward the wrong thing but this actually leads to behavior that appears aligned based on RL performance in distribution. 

We similarly visualize traces for each internal node in the sophisticated reward DDT trained for the Beam Rider game without penalty (Fig~\ref{fig:BR_wo_penalty}) and compare it against that of reward DDT trained using penalty (Fig~\ref{fig:BR_penalty}). In Fig~\ref{fig:BR_wo_penalty} we see that Node A routes states where agent hits an enemy ship to the left and states where it misses enemy ships to right. Then Node B routes states where it looks like it will hit an enemy ship to a reward of 1.0 but interestingly routes states where it has hit an enemy ship to a reward of 0 (yellow flash indicates an enemy being destroyed). This allows us to see a misalignment in the learned reward function. We investigated this further and found that when the agent loses a life, this also triggers a flashing yellow screen. Thus, the agent appears to be misinterpreting the yellow flash and associating it with losing a life, when it should be associated with a good reward for destroying an enemy ship. 
We also created traces for reward DDT trained with regularization penalty on routing probabilities for Breakout and BeamRider. We observe similar trends of misalignment (Node B) of the learned regularized reward DDT for BeamRider (Fig~\ref{fig:BR_penalty}).



In Table~\ref{tab:rl-results} we summarize learned policy performance under 4 different scenarios (without penalty and without argmax (returning soft reward averaged over all leaf nodes), without penalty and with argmax, with penalty and without argmax, with penalty and with argmax) for both Beam Rider and Breakout along with T-REX performance on each of these games. Our results in terms of performance are mixed. We find evidence that using the soft reward output ($\neg$argmax) of a DDT leads to the best RL performance.
Interestingly, we observe that penalty regularization helps RL performance in case of Breakout but leads to degradation in RL performance in case of Beam Rider. However, in terms of IQM, the RL performance when optimizing the DDT rewards is not able to match the performance of the end-to-end neural network baseline.

For Beam Rider, we examined the learned RL policies for both reward DDTs and TREX and found that the misaligned reward did lead to misaligned behavior: agents across various seeds for both DDT and TREX move to one end of a screen, learn to stay alive, but never fire at the enemy ship, thus avoiding getting hit but also avoiding scoring points. 
Notably, in case of reward DDTs, both with and without penalty, we could detect this misalignment before running RL using our synthetic traces, but for TREX which use a dense black box neural network for learning reward, we could not diagnose this misalignment in reward function prior to running RL.

\section{Discussion and Future Work}
Our work provides mixed results regarding the utility of reward DDTs. 
On one hand, we provide evidence that reward DDTs are a viable alternative to end-to-end deep network rewards and can sometimes perform on-par with their deep neural network reward counterparts; however, for complex domains like Atari, the best performance comes at the cost of using  DDT in a way that is not interpretable: using a soft reward output that is a weighted sum of outputs of all leaf nodes. Ideally, we could use reward DDTs with hard (argmax) reward outputs---the reward output during policy optimization would come from a single leaf node, allowing us to trace the reward output to a small number of binary routing decisions at the internal nodes. While optimizing this kind of hard output (argmax) process works well for the simpler domains we studied (e.g., CartPole and MNIST Gridworlds); it seems to hurt performance on more complex domains. We hypothesize this might be a result of the reward function being too sparse. Thus, our results reveal a tension between wanting highly shaped rewards to ensure good RL performance, while also wanting simple, non-shaped rewards to afford interpretability. Future work should investigate this trade-off in more depth.

In terms of interpretability, we find that for low dimensional tasks such as Cartpole and MNIST GridWorld environments, our framework is capable of providing global explanations that reveal interesting insights into the learned reward. For higher dimensional tasks such as Atari, we approximate global explanations by leveraging aggregations of local explanations by finding the input states that maximally and minimally activate the routing probability of each internal node. 
However, we also find that the deeper the DDT, the harder it becomes to interpret the learned reward. 
We also present evidence that demonstrates the practicality of using reward DDTs as a kind of \textit{alignment debugger tool} to inspect learned reward functions for alignment with human intent. In particular, we provide evidence that reward DDTs can reveal cases of silent misalignment. By running policy optimization, we also find that baseline black-box neural network rewards are also misaligned. Importantly, the interpretability of a reward DDT reveals the silent misalignment without needing to run RL. 
Future work should investigate using our framework to understand and interpret existing pre-trained neural network reward models that are known to lead to unintended consequences~\citep{christiano2017deep,ibarz2018reward,brown2021pg-broil,tien2023study} by distilling these networks into reward DDTs. Future work also includes investigating how to fix a known misaligned reward DDT by fine-tuning leaf and internal nodes based on human feedback, perhaps by using human-in-the-loop representation and feature learning~\citep{bobu2022inducing,bobu2023sirl} or using methods for identifying causal features using small amounts of human annotations~\citep{ghosal2023contextual}. Future work shouldalso explore how to extend the ideas in this paper to transformer-based reward functions used in LLMs~\citep{ouyang2022training} and investigate the effects of increasing tree depth on interpretability.

\subsubsection*{Acknowledgments}
\label{sec:ack}
This work has taken place in the Aligned, Robust, and Interactive Autonomy (ARIA) Lab at The University of Utah. ARIA Lab research is supported in part by the NSF (IIS-2310759), the NIH (R21EB035378), Open Philanthropy, and the ARL STRONG program.

{
\bibliographystyle{plainnat}
\bibliography{DDT}

\begin{thebibliography}{69}
\providecommand{\natexlab}[1]{#1}
\providecommand{\url}[1]{\texttt{#1}}
\expandafter\ifx\csname urlstyle\endcsname\relax
  \providecommand{\doi}[1]{doi: #1}\else
  \providecommand{\doi}{doi: \begingroup \urlstyle{rm}\Url}\fi

\bibitem[Agarwal et~al.(2021)Agarwal, Schwarzer, Castro, Courville, and Bellemare]{agarwal2021deep}
Rishabh Agarwal, Max Schwarzer, Pablo~Samuel Castro, Aaron~C Courville, and Marc Bellemare.
\newblock Deep reinforcement learning at the edge of the statistical precipice.
\newblock \emph{Advances in neural information processing systems}, 34:\penalty0 29304--29320, 2021.

\bibitem[{Bellemare} et~al.(2013){Bellemare}, {Naddaf}, {Veness}, and {Bowling}]{ale}
M.~G. {Bellemare}, Y.~{Naddaf}, J.~{Veness}, and M.~{Bowling}.
\newblock The arcade learning environment: An evaluation platform for general agents.
\newblock \emph{Journal of Artificial Intelligence Research}, 47:\penalty0 253--279, Jun 2013.

\bibitem[Bellemare et~al.(2013)Bellemare, Naddaf, Veness, and Bowling]{bellemare2013arcade}
Marc~G Bellemare, Yavar Naddaf, Joel Veness, and Michael Bowling.
\newblock The arcade learning environment: An evaluation platform for general agents.
\newblock \emph{Journal of Artificial Intelligence Research}, 47:\penalty0 253--279, 2013.

\bibitem[Bewley and Lecue(2022)]{bewley2022interpretable}
Tom Bewley and Freddy Lecue.
\newblock Interpretable preference-based reinforcement learning with tree-structured reward functions.
\newblock In \emph{Proceedings of the 21st International Conference on Autonomous Agents and Multiagent Systems}, pages 118--126, 2022.

\bibitem[Bewley et~al.(2023)Bewley, Lawry, Richards, Craddock, and Henderson]{bewley2023reward}
Tom Bewley, Jonathan Lawry, Arthur Richards, Rachel Craddock, and Ian Henderson.
\newblock Reward learning with trees: Methods and evaluation, 2023.
\newblock URL \url{https://openreview.net/forum?id=xl2-MIX2DCD}.

\bibitem[Biyik et~al.(2020)Biyik, Palan, Landolfi, Losey, Sadigh, et~al.]{erdem2020asking}
Erdem Biyik, Malayandi Palan, Nicholas~C Landolfi, Dylan~P Losey, Dorsa Sadigh, et~al.
\newblock Asking easy questions: A user-friendly approach to active reward learning.
\newblock In \emph{Conference on Robot Learning}, pages 1177--1190. PMLR, 2020.

\bibitem[Bobu et~al.(2022)Bobu, Wiggert, Tomlin, and Dragan]{bobu2022inducing}
Andreea Bobu, Marius Wiggert, Claire Tomlin, and Anca~D Dragan.
\newblock Inducing structure in reward learning by learning features.
\newblock \emph{The International Journal of Robotics Research}, 41\penalty0 (5):\penalty0 497--518, 2022.

\bibitem[Bobu et~al.(2023)Bobu, Liu, Shah, Brown, and Dragan]{bobu2023sirl}
Andreea Bobu, Yi~Liu, Rohin Shah, Daniel~S. Brown, and Anca~D. Dragan.
\newblock Sirl: Similarity-based implicit representation learning.
\newblock In \emph{Proceedings of the 2023 ACM/IEEE International Conference on Human-Robot Interaction (HRI)}, 2023.

\bibitem[Booth et~al.(2022)Booth, Sharma, Chung, Shah, and Glassman]{booth2022revisiting}
Serena Booth, Sanjana Sharma, Sarah Chung, Julie Shah, and Elena~L Glassman.
\newblock Revisiting human-robot teaching and learning through the lens of human concept learning.
\newblock In \emph{2022 17th ACM/IEEE International Conference on Human-Robot Interaction (HRI)}, pages 147--156. IEEE, 2022.

\bibitem[Bradley and Terry(1952)]{bradley1952rank}
Ralph~Allan Bradley and Milton~E Terry.
\newblock Rank analysis of incomplete block designs: I. the method of paired comparisons.
\newblock \emph{Biometrika}, 39\penalty0 (3/4):\penalty0 324--345, 1952.

\bibitem[Brockman et~al.(2016)Brockman, Cheung, Pettersson, Schneider, Schulman, Tang, and Zaremba]{brockman2016openai}
Greg Brockman, Vicki Cheung, Ludwig Pettersson, Jonas Schneider, John Schulman, Jie Tang, and Wojciech Zaremba.
\newblock Openai gym.
\newblock \emph{arXiv preprint arXiv:1606.01540}, 2016.

\bibitem[Brown et~al.(2019)Brown, Goo, Nagarajan, and Niekum]{brown2019extrapolating}
Daniel~S. Brown, Wonjoon Goo, Prabhat Nagarajan, and Scott Niekum.
\newblock Extrapolating beyond suboptimal demonstrations via inverse reinforcement learning from observations.
\newblock In \emph{International conference on machine learning}, pages 783--792. PMLR, 2019.

\bibitem[Brown et~al.(2020)Brown, Goo, and Niekum]{brown2020better}
Daniel~S. Brown, Wonjoon Goo, and Scott Niekum.
\newblock Better-than-demonstrator imitation learning via automatically-ranked demonstrations.
\newblock In \emph{Conference on robot learning}, pages 330--359. PMLR, 2020.

\bibitem[Brown et~al.(2021)Brown, Schneider, Dragan, and Niekum]{brown2021value}
Daniel~S. Brown, Jordan Schneider, Anca Dragan, and Scott Niekum.
\newblock Value alignment verification.
\newblock In \emph{International Conference on Machine Learning}, pages 1105--1115. PMLR, 2021.

\bibitem[Christiano et~al.(2017)Christiano, Leike, Brown, Martic, Legg, and Amodei]{christiano2017deep}
Paul~F Christiano, Jan Leike, Tom~B Brown, Miljan Martic, Shane Legg, and Dario Amodei.
\newblock Deep reinforcement learning from human preferences.
\newblock In \emph{NIPS}, 2017.

\bibitem[Coppens et~al.(2019)Coppens, Efthymiadis, Lenaerts, Now{\'e}, Miller, Weber, and Magazzeni]{coppens2019distilling}
Youri Coppens, Kyriakos Efthymiadis, Tom Lenaerts, Ann Now{\'e}, Tim Miller, Rosina Weber, and Daniele Magazzeni.
\newblock Distilling deep reinforcement learning policies in soft decision trees.
\newblock In \emph{Proceedings of the IJCAI 2019 workshop on explainable artificial intelligence}, pages 1--6, 2019.

\bibitem[Devidze et~al.(2021)Devidze, Radanovic, Kamalaruban, and Singla]{devidze2021explicable}
Rati Devidze, Goran Radanovic, Parameswaran Kamalaruban, and Adish Singla.
\newblock Explicable reward design for reinforcement learning agents.
\newblock \emph{Advances in Neural Information Processing Systems}, 34:\penalty0 20118--20131, 2021.

\bibitem[Ding et~al.(2021)Ding, Hernandez-Leal, Ding, Li, and Huang]{ding2021cdt}
Zihan Ding, Pablo Hernandez-Leal, Gavin~Weiguang Ding, Changjian Li, and Ruitong Huang.
\newblock Cdt: Cascading decision trees for explainable reinforcement learning, 2021.

\bibitem[Finn et~al.(2016)Finn, Levine, and Abbeel]{finn2016guided}
Chelsea Finn, Sergey Levine, and Pieter Abbeel.
\newblock Guided cost learning: Deep inverse optimal control via policy optimization.
\newblock In \emph{International conference on machine learning}, pages 49--58. PMLR, 2016.

\bibitem[Fisac et~al.(2020)Fisac, Gates, Hamrick, Liu, Hadfield-Menell, Palaniappan, Malik, Sastry, Griffiths, and Dragan]{fisac2020pragmatic}
Jaime~F Fisac, Monica~A Gates, Jessica~B Hamrick, Chang Liu, Dylan Hadfield-Menell, Malayandi Palaniappan, Dhruv Malik, S~Shankar Sastry, Thomas~L Griffiths, and Anca~D Dragan.
\newblock Pragmatic-pedagogic value alignment.
\newblock In \emph{Robotics Research: The 18th International Symposium ISRR}, pages 49--57. Springer, 2020.

\bibitem[Freire et~al.(2020)Freire, Gleave, Toyer, and Russell]{freire2020derail}
Pedro Freire, Adam Gleave, Sam Toyer, and Stuart Russell.
\newblock Derail: Diagnostic environments for reward and imitation learning.
\newblock \emph{arXiv preprint arXiv:2012.01365}, 2020.

\bibitem[Frosst and Hinton(2017)]{frosst2017distilling}
Nicholas Frosst and Geoffrey Hinton.
\newblock Distilling a neural network into a soft decision tree.
\newblock \emph{arXiv preprint arXiv:1711.09784}, 2017.

\bibitem[Fu et~al.(2018)Fu, Luo, and Levine]{fu2018learning}
Justin Fu, Katie Luo, and Sergey Levine.
\newblock Learning robust rewards with adverserial inverse reinforcement learning.
\newblock In \emph{International Conference on Learning Representations}, 2018.

\bibitem[Ghosal et~al.(2023{\natexlab{a}})Ghosal, Zurek, Brown, and Dragan]{ghosal2022effect}
Gaurav~R Ghosal, Matthew Zurek, Daniel~S Brown, and Anca~D Dragan.
\newblock The effect of modeling human rationality level on learning rewards from multiple feedback types.
\newblock \emph{AAAI Conference on Artificial Intelligence}, 2023{\natexlab{a}}.

\bibitem[Ghosal et~al.(2023{\natexlab{b}})Ghosal, Setlur, Brown, Dragan, and Raghunathan]{ghosal2023contextual}
Gaurav~Rohit Ghosal, Amrith Setlur, Daniel~S. Brown, Anca Dragan, and Aditi Raghunathan.
\newblock Contextual reliability: When different features matter in different contexts.
\newblock In \emph{International Conference on Machine Learning (ICML)}, 2023{\natexlab{b}}.

\bibitem[Gilpin et~al.(2018)Gilpin, Bau, Yuan, Bajwa, Specter, and Kagal]{gilpin2018explaining}
Leilani~H Gilpin, David Bau, Ben~Z Yuan, Ayesha Bajwa, Michael Specter, and Lalana Kagal.
\newblock Explaining explanations: An overview of interpretability of machine learning.
\newblock In \emph{2018 IEEE 5th International Conference on data science and advanced analytics (DSAA)}, pages 80--89. IEEE, 2018.

\bibitem[Gleave et~al.(2021)Gleave, Dennis, Legg, Russell, and Leike]{gleave2020quantifying}
Adam Gleave, Michael Dennis, Shane Legg, Stuart Russell, and Jan Leike.
\newblock Quantifying differences in reward functions.
\newblock In \emph{International Conference on Learning Representations}, 2021.
\newblock URL \url{https://openreview.net/forum?id=LwEQnp6CYev}.

\bibitem[Hazimeh et~al.(2020)Hazimeh, Ponomareva, Mol, Tan, and Mazumder]{hazimeh2020tree}
Hussein Hazimeh, Natalia Ponomareva, Petros Mol, Zhenyu Tan, and Rahul Mazumder.
\newblock The tree ensemble layer: Differentiability meets conditional computation.
\newblock In \emph{International Conference on Machine Learning}, pages 4138--4148. PMLR, 2020.

\bibitem[Hejna and Sadigh(2022)]{henja2022fewshot}
Donald~Joseph Hejna and Dorsa Sadigh.
\newblock Few-shot preference learning for human-in-the-loop {RL}.
\newblock In \emph{6th Annual Conference on Robot Learning}, 2022.
\newblock URL \url{https://openreview.net/forum?id=IKC5TfXLuW0}.

\bibitem[Heuillet et~al.(2021)Heuillet, Couthouis, and D{\'\i}az-Rodr{\'\i}guez]{heuillet2021explainability}
Alexandre Heuillet, Fabien Couthouis, and Natalia D{\'\i}az-Rodr{\'\i}guez.
\newblock Explainability in deep reinforcement learning.
\newblock \emph{Knowledge-Based Systems}, 214:\penalty0 106685, 2021.

\bibitem[Ibarz et~al.(2018)Ibarz, Leike, Pohlen, Irving, Legg, and Amodei]{ibarz2018reward}
Borja Ibarz, Jan Leike, Tobias Pohlen, Geoffrey Irving, Shane Legg, and Dario Amodei.
\newblock Reward learning from human preferences and demonstrations in atari.
\newblock \emph{arXiv preprint arXiv:1811.06521}, 2018.

\bibitem[Icarte et~al.(2022)Icarte, Klassen, Valenzano, and McIlraith]{icarte2022reward}
Rodrigo~Toro Icarte, Toryn~Q Klassen, Richard Valenzano, and Sheila~A McIlraith.
\newblock Reward machines: Exploiting reward function structure in reinforcement learning.
\newblock \emph{Journal of Artificial Intelligence Research}, 73:\penalty0 173--208, 2022.

\bibitem[Javed et~al.(2021)Javed, Brown, Sharma, Zhu, Balakrishna, Petrik, Dragan, and Goldberg]{brown2021pg-broil}
Zaynah Javed, Daniel~S. Brown, Satvik Sharma, Jerry Zhu, Ashwin Balakrishna, Marek Petrik, Anca~D. Dragan, and Ken Goldberg.
\newblock Policy gradient bayesian robust optimization.
\newblock In \emph{International Conference on Machine Learning (ICML)}, 2021.

\bibitem[Jeon et~al.(2020)Jeon, Milli, and Dragan]{jeon2020reward}
Hong~Jun Jeon, Smitha Milli, and Anca Dragan.
\newblock Reward-rational (implicit) choice: A unifying formalism for reward learning.
\newblock \emph{Advances in Neural Information Processing Systems}, 33:\penalty0 4415--4426, 2020.

\bibitem[Ji et~al.(2023)Ji, Qiu, Chen, Zhang, Lou, Wang, Duan, He, Zhou, Zhang, et~al.]{ji2023ai}
Jiaming Ji, Tianyi Qiu, Boyuan Chen, Borong Zhang, Hantao Lou, Kaile Wang, Yawen Duan, Zhonghao He, Jiayi Zhou, Zhaowei Zhang, et~al.
\newblock Ai alignment: A comprehensive survey.
\newblock \emph{arXiv preprint arXiv:2310.19852}, 2023.

\bibitem[Jiang et~al.(2021)Jiang, Bharadwaj, Wu, Shah, Topcu, and Stone]{jiang2021temporal}
Yuqian Jiang, Suda Bharadwaj, Bo~Wu, Rishi Shah, Ufuk Topcu, and Peter Stone.
\newblock Temporal-logic-based reward shaping for continuing reinforcement learning tasks.
\newblock In \emph{Proceedings of the AAAI Conference on Artificial Intelligence}, volume~35, pages 7995--8003, 2021.

\bibitem[Jordan(1994)]{jordanDT}
Michael~I. Jordan.
\newblock A statistical approach to decision tree modeling.
\newblock In \emph{Proceedings of the Eleventh International Conference on International Conference on Machine Learning}, ICML'94, page 363–370, San Francisco, CA, USA, 1994. Morgan Kaufmann Publishers Inc.
\newblock ISBN 1558603352.

\bibitem[Karimi et~al.(2024)Karimi, Ho, Thach, Kuntz, and Brown]{karimi2024reward}
Zohre Karimi, Shing-Hei Ho, Bao Thach, Alan Kuntz, and Daniel~S Brown.
\newblock Reward learning from suboptimal demonstrations with applications in surgical electrocautery.
\newblock \emph{International Symposium on Medical Robotics (ISMR)}, 2024.

\bibitem[Kingma and Ba(2014)]{kingma2014adam}
Diederik~P Kingma and Jimmy Ba.
\newblock Adam: A method for stochastic optimization.
\newblock \emph{arXiv preprint arXiv:1412.6980}, 2014.

\bibitem[Kotsiantis(2013)]{kotsiantis2013decision}
Sotiris~B Kotsiantis.
\newblock Decision trees: a recent overview.
\newblock \emph{Artificial Intelligence Review}, 39:\penalty0 261--283, 2013.

\bibitem[Krakovna et~al.(2020)Krakovna, Uesato, Mikulik, Rahtz, Everitt, Kumar, Kenton, Leike, and Legg]{krakovna2020specification}
Victoria Krakovna, Jonathan Uesato, Vladimir Mikulik, Matthew Rahtz, Tom Everitt, Ramana Kumar, Zac Kenton, Jan Leike, and Shane Legg.
\newblock Specification gaming: the flip side of ai ingenuity.
\newblock \emph{DeepMind Blog}, 2020.

\bibitem[Lee et~al.(2021{\natexlab{a}})Lee, Smith, and Abbeel]{lee2021pebble}
Kimin Lee, Laura Smith, and Pieter Abbeel.
\newblock Pebble: Feedback-efficient interactive reinforcement learning via relabeling experience and unsupervised pre-training.
\newblock \emph{arXiv preprint arXiv:2106.05091}, 2021{\natexlab{a}}.

\bibitem[Lee et~al.(2021{\natexlab{b}})Lee, Admoni, and Simmons]{lee2021machine}
Michael~S Lee, Henny Admoni, and Reid Simmons.
\newblock Machine teaching for human inverse reinforcement learning.
\newblock \emph{Frontiers in Robotics and AI}, 8:\penalty0 693050, 2021{\natexlab{b}}.

\bibitem[Leike et~al.(2018)Leike, Krueger, Everitt, Martic, Maini, and Legg]{leike2018scalable}
Jan Leike, David Krueger, Tom Everitt, Miljan Martic, Vishal Maini, and Shane Legg.
\newblock Scalable agent alignment via reward modeling: a research direction.
\newblock \emph{arXiv preprint arXiv:1811.07871}, 2018.

\bibitem[Liu et~al.(2023)Liu, Datta, Novoseller, and Brown]{liu2023efficient}
Yi~Liu, Gaurav Datta, Ellen Novoseller, and Daniel~S Brown.
\newblock Efficient preference-based reinforcement learning using learned dynamics models.
\newblock In \emph{2023 IEEE International Conference on Robotics and Automation (ICRA)}, pages 2921--2928. IEEE, 2023.

\bibitem[Mahmud et~al.(2023)Mahmud, Saisubramanian, and Zilberstein]{mahmud2023reveale}
Saaduddin Mahmud, Sandhya Saisubramanian, and Shlomo Zilberstein.
\newblock Reveale: Reward verification and learning using explanations.
\newblock 2023.

\bibitem[Mehta and Losey(2022)]{mehta2022unified}
Shaunak~A Mehta and Dylan~P Losey.
\newblock Unified learning from demonstrations, corrections, and preferences during physical human-robot interaction.
\newblock \emph{arXiv preprint arXiv:2207.03395}, 2022.

\bibitem[Michaud et~al.(2020)Michaud, Gleave, and Russell]{michaud2020understanding}
Eric~J Michaud, Adam Gleave, and Stuart Russell.
\newblock Understanding learned reward functions.
\newblock \emph{arXiv preprint arXiv:2012.05862}, 2020.

\bibitem[Mnih et~al.(2016)Mnih, Badia, Mirza, Graves, Lillicrap, Harley, Silver, and Kavukcuoglu]{mnih2016asynchronous}
Volodymyr Mnih, Adria~Puigdomenech Badia, Mehdi Mirza, Alex Graves, Timothy Lillicrap, Tim Harley, David Silver, and Koray Kavukcuoglu.
\newblock Asynchronous methods for deep reinforcement learning.
\newblock In \emph{International conference on machine learning}, pages 1928--1937. PMLR, 2016.

\bibitem[Molnar(2020)]{molnar2020interpretable}
Christoph Molnar.
\newblock \emph{Interpretable machine learning}.
\newblock Lulu. com, 2020.

\bibitem[Ng et~al.(1999)Ng, Harada, and Russell]{ng1999policy}
Andrew~Y Ng, Daishi Harada, and Stuart Russell.
\newblock Policy invariance under reward transformations: Theory and application to reward shaping.
\newblock In \emph{Icml}, volume~99, pages 278--287, 1999.

\bibitem[Ouyang et~al.(2022)Ouyang, Wu, Jiang, Almeida, Wainwright, Mishkin, Zhang, Agarwal, Slama, Ray, et~al.]{ouyang2022training}
Long Ouyang, Jeffrey Wu, Xu~Jiang, Diogo Almeida, Carroll Wainwright, Pamela Mishkin, Chong Zhang, Sandhini Agarwal, Katarina Slama, Alex Ray, et~al.
\newblock Training language models to follow instructions with human feedback.
\newblock \emph{Advances in Neural Information Processing Systems}, 35:\penalty0 27730--27744, 2022.

\bibitem[Pace et~al.(2022)Pace, Chan, and van~der Schaar]{pace2022poetree}
Alizée Pace, Alex~J. Chan, and Mihaela van~der Schaar.
\newblock Poetree: Interpretable policy learning with adaptive decision trees, 2022.

\bibitem[Patterson et~al.(2023)Patterson, Neumann, White, and White]{patterson2023empirical}
Andrew Patterson, Samuel Neumann, Martha White, and Adam White.
\newblock Empirical design in reinforcement learning.
\newblock \emph{arXiv preprint arXiv:2304.01315}, 2023.

\bibitem[Quinlan(1986)]{quinlan1986induction}
J.~Ross Quinlan.
\newblock Induction of decision trees.
\newblock \emph{Machine learning}, 1:\penalty0 81--106, 1986.

\bibitem[R{\"a}ukur et~al.(2022)R{\"a}ukur, Ho, Casper, and Hadfield-Menell]{raukur2022toward}
Tilman R{\"a}ukur, Anson Ho, Stephen Casper, and Dylan Hadfield-Menell.
\newblock Toward transparent ai: A survey on interpreting the inner structures of deep neural networks.
\newblock \emph{arXiv preprint arXiv:2207.13243}, 2022.

\bibitem[Russell et~al.(2015)Russell, Dewey, and Tegmark]{russell2015research}
Stuart Russell, Daniel Dewey, and Max Tegmark.
\newblock Research priorities for robust and beneficial artificial intelligence.
\newblock \emph{Ai Magazine}, 36\penalty0 (4):\penalty0 105--114, 2015.

\bibitem[Sadigh et~al.(2017)Sadigh, Dragan, Sastry, and Seshia]{sadigh2017active}
Dorsa Sadigh, Anca~D Dragan, Shankar Sastry, and Sanjit~A Seshia.
\newblock Active preference-based learning of reward functions.
\newblock In \emph{Robotics: Science and Systems}, 2017.

\bibitem[Sanneman and Shah(2022)]{9822391}
Lindsay Sanneman and Julie~A. Shah.
\newblock An empirical study of reward explanations with human-robot interaction applications.
\newblock \emph{IEEE Robotics and Automation Letters}, 7\penalty0 (4):\penalty0 8956--8963, 2022.
\newblock \doi{10.1109/LRA.2022.3189441}.

\bibitem[Silva et~al.(2020)Silva, Gombolay, Killian, Jimenez, and Son]{silva2020optimization}
Andrew Silva, Matthew Gombolay, Taylor Killian, Ivan Jimenez, and Sung-Hyun Son.
\newblock Optimization methods for interpretable differentiable decision trees applied to reinforcement learning.
\newblock In \emph{International conference on artificial intelligence and statistics}, pages 1855--1865. PMLR, 2020.

\bibitem[Sutton and Barto(2018)]{sutton2018reinforcement}
Richard~S Sutton and Andrew~G Barto.
\newblock \emph{Reinforcement learning: An introduction}.
\newblock MIT press, 2018.

\bibitem[Tambwekar et~al.(2023)Tambwekar, Silva, Gopalan, and Gombolay]{10105980}
Pradyumna Tambwekar, Andrew Silva, Nakul Gopalan, and Matthew Gombolay.
\newblock Natural language specification of reinforcement learning policies through differentiable decision trees.
\newblock \emph{IEEE Robotics and Automation Letters}, pages 1--8, 2023.
\newblock \doi{10.1109/LRA.2023.3268593}.

\bibitem[Tanno et~al.(2019)Tanno, Arulkumaran, Alexander, Criminisi, and Nori]{tanno2019adaptive}
Ryutaro Tanno, Kai Arulkumaran, Daniel Alexander, Antonio Criminisi, and Aditya Nori.
\newblock Adaptive neural trees.
\newblock In \emph{International Conference on Machine Learning}, pages 6166--6175. PMLR, 2019.

\bibitem[Tien et~al.(2023)Tien, He, Erickson, Dragan, and Brown]{tien2023study}
Jeremy Tien, Jerry Zhi-Yang He, Zackory Erickson, Anca Dragan, and Daniel~S Brown.
\newblock Causal confusion and reward misidentification in preference-based reward learning.
\newblock In \emph{International Conference on Learning Representations}, 2023.

\bibitem[Tucker et~al.(2018)Tucker, Gleave, and Russell]{tucker2018inverse}
Aaron Tucker, Adam Gleave, and Stuart Russell.
\newblock Inverse reinforcement learning for video games.
\newblock \emph{arXiv preprint arXiv:1810.10593}, 2018.

\bibitem[Wirth et~al.(2016)Wirth, F{\"u}rnkranz, and Neumann]{wirth2016model}
Christian Wirth, Johannes F{\"u}rnkranz, and Gerhard Neumann.
\newblock Model-free preference-based reinforcement learning.
\newblock In \emph{Thirtieth AAAI Conference on Artificial Intelligence}, 2016.

\bibitem[Zakka et~al.(2022)Zakka, Zeng, Florence, Tompson, Bohg, and Dwibedi]{zakka2022xirl}
Kevin Zakka, Andy Zeng, Pete Florence, Jonathan Tompson, Jeannette Bohg, and Debidatta Dwibedi.
\newblock Xirl: Cross-embodiment inverse reinforcement learning.
\newblock In \emph{Conference on Robot Learning}, pages 537--546. PMLR, 2022.

\bibitem[Zantedeschi et~al.(2021)Zantedeschi, Kusner, and Niculae]{zantedeschi2021learning}
Valentina Zantedeschi, Matt~J Kusner, and Vlad Niculae.
\newblock Learning binary trees by argmin differentiation.
\newblock \emph{ICML}, 2021.

\bibitem[Zhang and Zhu(2018)]{zhang2018visual}
Quan-shi Zhang and Song-Chun Zhu.
\newblock Visual interpretability for deep learning: a survey.
\newblock \emph{Frontiers of Information Technology \& Electronic Engineering}, 19\penalty0 (1):\penalty0 27--39, 2018.

\end{thebibliography}
}
\appendix
\newpage

\section{DDT Routing Penalty Regularization}\label{app:A_penalty}
We take inspiration from [19] for adding penalty regularization and we first explain how penalty is defined at each internal node and then elaborate on calculating penalty for a single state over the whole DDT.

The cross-entropy between desired routing probability distribution of an internal node such that it's children nodes are equally used and the actual routing probability distribution is referred to as Penalty and is given by

\begin{equation}
\alpha_i=\frac{\sum_{\mathbf{x}} P^i(\mathbf{x}) p_i(\mathbf{x})}{\sum_{\mathbf{x}} P^i(\mathbf{x})}
\end{equation}

where the probability of a current internal node is $p_{i}(\textbf{x})$ and path probability from root node to an internal node is $P^{i}(\textbf{x})$.

Penalty over the whole DDT for a single state is defined as sum over all internal nodes for the given input \textbf{x}

\begin{equation}
C=-\lambda \sum_{i \in \text { Inner Nodes }} 0.5 \log \left(\alpha_i\right)+0.5 \log \left(1-\alpha_i\right)
\end{equation}
 where hyper-parameter $\lambda$  controls the strength of penalty $\lambda$ in reward DDT so that the penalty strength is proportional to $2^{-d}$ and decays  exponentially with depth of tree.
Finally the penalty term  for learning reward tree from pairwise preferences is calculated by taking the mean over all penalties for all states in the pairwise demonstrations.

\section{Cartpole} \label{app:cartpole}
The baseline neural network is comprised of 2 fully connected layers, each of dimension 16 to learn the reward function. For reward models, both DDT and neural network, we use  2000 training and 200 validation pairwise preference demonstrations, each of length 20, with Adam optimizer and $lr=0.001$ and weight decay$=0$. 

For running RL on ground truth reward as well as under learned reward models, we use Stable Baselines3 PPO with $batch size=1024,lr=0.001,gaelambda=0.8, gamma =0.98, nepochs=20, nsteps=2048$ for 500000 total timesteps across 5 environments for both In-Distribution and Out-Of-Distribution starting cart positions.

In case of In-Distribution starting cart positions, we found that out of 10 seeds that we report results on, 3 seeds of RL policy learned under the neural network reward function seem to suffer from catastrophic forgetting leading a high standard deviation , with Mean and IQM, marginally lower than RL performance under ground truth reward function as well as performance of policices learned under soft reward and maximum probability path of DDT. 

\section{MNIST Gridworld Additional Details}\label{app:mnist_val}
 In this environment,the action space $a$ contains $4$ main actions: go left, go right, move up, move down. The transition function is stochastic and moves the agent in the direction chosen  with an $80\%$ probability as long as the action does not take it off of the grid. Actions that would result in leaving the grid result in a self transition. 

 And the neural network used to learn reward from pairwise human preferences  consisted 2 convolutional layers with kernel size 7 and 5 respectively and stride 1 with LeakyRelu as the non-linearities  followed by 2 fully connected layers. 

In Table ~\ref{MNIST-ValPrefAccuracy-table}, we report the accuracy of learned reward DDTs with CRL and IL leaf nodes over pairwise preferences generated using the held-out validation dataset,  both when using soft reward outputs and when using the output of the maximum probability path in the tree
(argmax) for both type of our DDT and compare it against that of a convolutional neural network. Our results show that our IL DDTs can often achieve high accuracy despite not using any convolutional filters even on held out data. Using soft reward with IL leaf nodes offers comparable accuracy to that of CNN while using the reward from maximum probability path leads to a small decrease in accuracy, but still performs better than a DDT with CRL leaf nodes in both soft and argmax settings. 

 \begin{table*}[h]
\footnotesize
  \caption{Accuracy of the learned reward models on the 25000 pairwise preferences generated using the held-out validation dataset. Since MNIST 0-3  Gridworld is  of size 5x5, we use trajectory length of 5 in the pairwise preferences while MNIST 0-9  Gridworld has size 10x10 , thus we use trajectory length 10.}
  \label{MNIST-ValPrefAccuracy-table}
  \centering
  \begin{tabular}{lrrrrr}
    \toprule
    &\multicolumn{4}{c}{Reward DDT} & \multicolumn{1}{c}{Baselines}                 \\
    & CRL Soft    & CRL Argmax   & IL Soft  & IL Argmax  & NNet\\
    \midrule
  
    MNIST 0-3  & 77.57\% & 77.10\% & 96.49\% & 94.10\% & 99.21\%  \\
    MNIST 0-9 & 72.71\% & 68.46\% & 90.83\% & 81.78\% & 92.40\%\\
    \bottomrule
  \end{tabular}
\end{table*}
 
\section{Additional domain: MNIST (0/1) Gridworld} \label{app:mnist01grid}

We also show here an even simpler version of MNIST gridworld where there are only two possible digits. For training the reward DDT with simple internal nodes and CRL leaf nodes, we use a learning rate of $0.001$, weight decay of $0.05$, and the Adam optimizer~\citep{kingma2014adam}.


\begin{figure}[t]
         \centering
     \begin{subfigure}[b]{0.3\linewidth}
         \centering
         \includegraphics[width=0.8\textwidth]{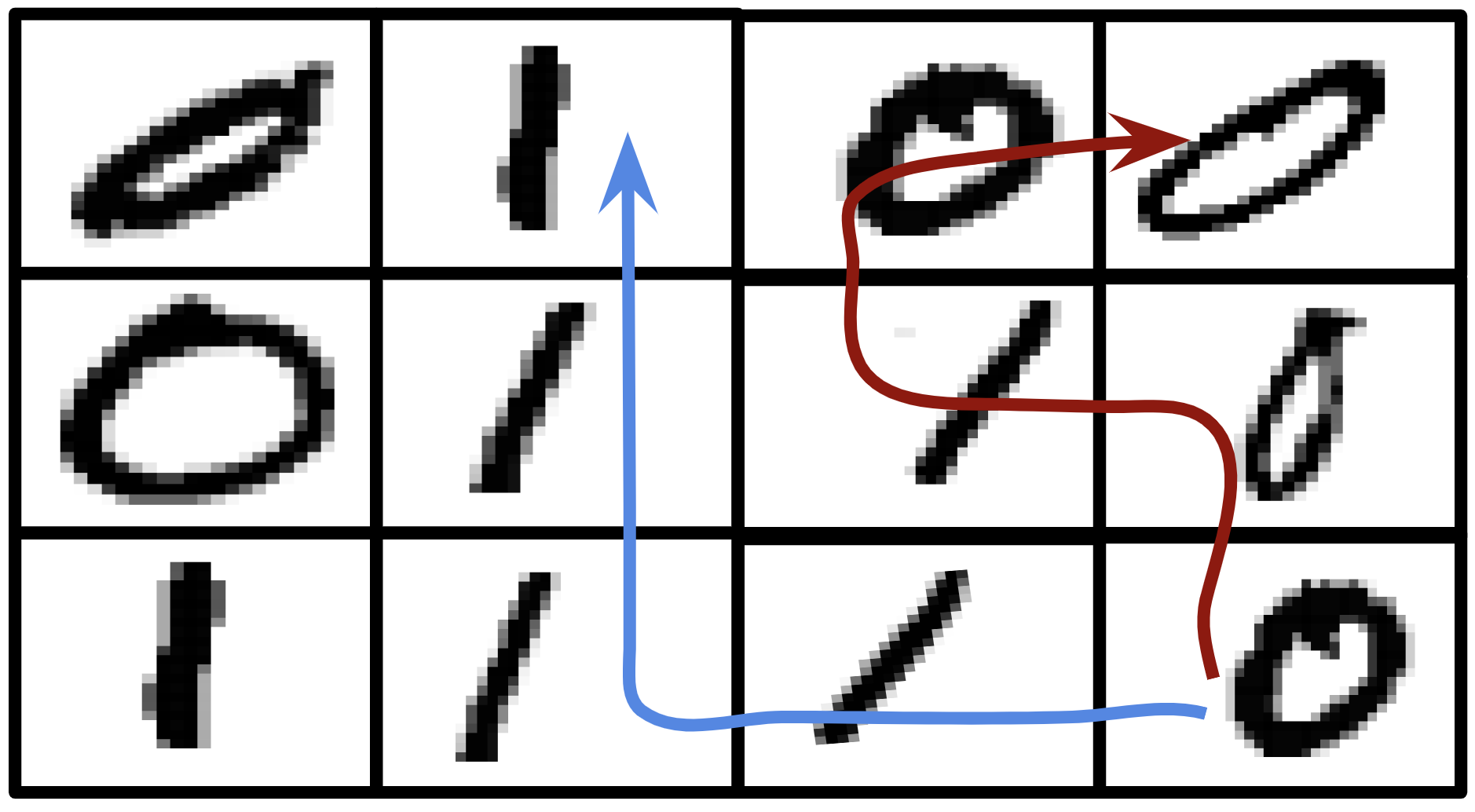}
         \caption{Pairwise trajectory preference}
         \label{fig:mnist_traj}
          \end{subfigure}
          \hspace{1cm}
      \begin{subfigure}[b]{0.4\linewidth}
         \centering
         \includegraphics[width=0.9\textwidth]{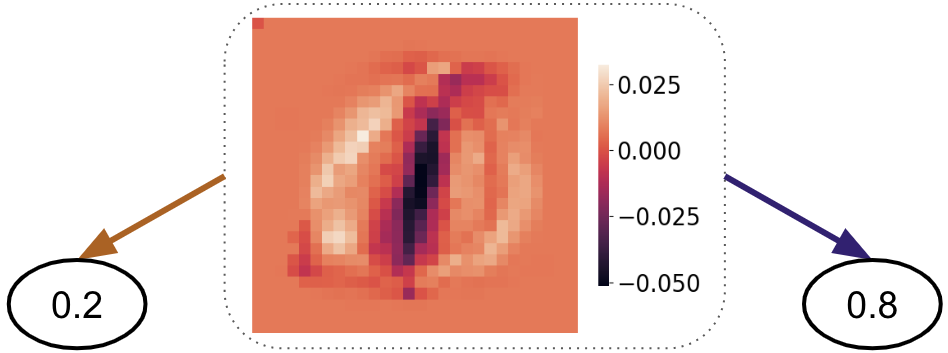}
         \caption{Visualization of Learned Reward DDT}
    \label{fig:bl}
    \end{subfigure}
     \hfill  
\caption{\textbf{MNIST (0/1) Gridworld.} (a) A pair of trajectories with the same starting state, where the blue trajectory (which visits more 1's) is preferred over the red trajectory.  (b) Heatmap of Learned Reward DDT : The dark pixels at center of heatmap form an approximate shape of digit 1 and are routed to right as the dark colors in heatmap mean that those pixels are turned off, while lighter pixels represent shape of digit 0 and routed to left as those pixels are turned on. Leaf nodes are depicted as circular nodes with their soft reward values.}
\label{fig:mnistbl}
\end{figure}

 \begin{table*}[h]
\footnotesize
  \caption{RL Performance as the percentage of expected return obtained relative to the performance of
an optimal policy on the ground-truth reward. }
  \label{MNIST01-RL-table}
  \centering
  \begin{tabular}{lrrrrrr}
    \toprule
    &\multicolumn{4}{c}{Reward DDT} & \multicolumn{2}{c}{Baselines}                 \\
    & CRL Soft    & CRL Argmax   & IL Soft  & IL Argmax  & NNet & Random\\
    \midrule
    MNIST 0-1   &92.37\%  & 82.27\%  & 99.98 & 100\% & 98.2\% & 7.38\%  \\

    \bottomrule
  \end{tabular}
\end{table*}
\paragraph{Setup}
We begin by examining our framework for image based inputs on the simplest gridworld environment.
In this 5x5 gridworld each state in the MDP corresponds to a MNIST digit 0 or 1. To test whether we can learn an interpretable reward function from pairs of preference demonstrations over trajectories (see Fig \ref{fig:mnist_traj} for an example), we modeled the reward as a DDT of depth 1 with one simple internal node as the root node and 2 CRL leaf nodes with reward vector $\textbf{R} = (0.0, 1.0)$. We also compare the RL performance of the same reward DDT but with IL leaf nodes.


\paragraph{Results}
The resulting heatmap in Fig~\ref{fig:bl} provides evidence that the reward DDT learns to branch based on visually interpretable features that correspond to a hand-written 0 (routes to left leaf node) and a hand-written 1 (routes to right leaf node). The RL performance using the Soft Reward from CRL Leaf DDT on MNIST 0-1 environment is shown in Table~\ref{MNIST01-RL-table} is comparable to a deep neural network reward function trained on pairwise preferences. We observe that taking the maximum probability path across the learned reward tree results in a small decrease in performance relative to when we take soft reward from the learned DDT. The RL performance of IL Leaf DDT outperforms that of both CRL DDT and a deep neural network, hence it provides evidence that our DDT framework is capable of learning an interpretable and useful reward function. 
      

\section{MNIST (0-3) Gridworld Additional Results and Analysis}\label{app:mnist_grid}
In this section we provide detailed analysis about interpretability of different DDTs, beginning from comparison between Reward DDT and Classification DDT, then comparing Reward DDTs constructed using two different leaf node formulations, followed by comparison of different regularization on a reward DDT. 

Note that for both reward DDTs with different leaf nodes CRL and IL, we trained using a learning rate of $0.001$ and weight decay $0.005$ and the Adam optimizer. And the neural network details are same as defined above in Appendix ~\ref{app:mnist01grid}.

\subsection{Min-Max Reward Interpolation Tree vs Classification tree}\label{app:minmax_vs_classification}

\begin{figure*}[h]
     \centering
     \begin{subfigure}[t]{0.85\linewidth}
         \centering
         \includegraphics[width=\textwidth]{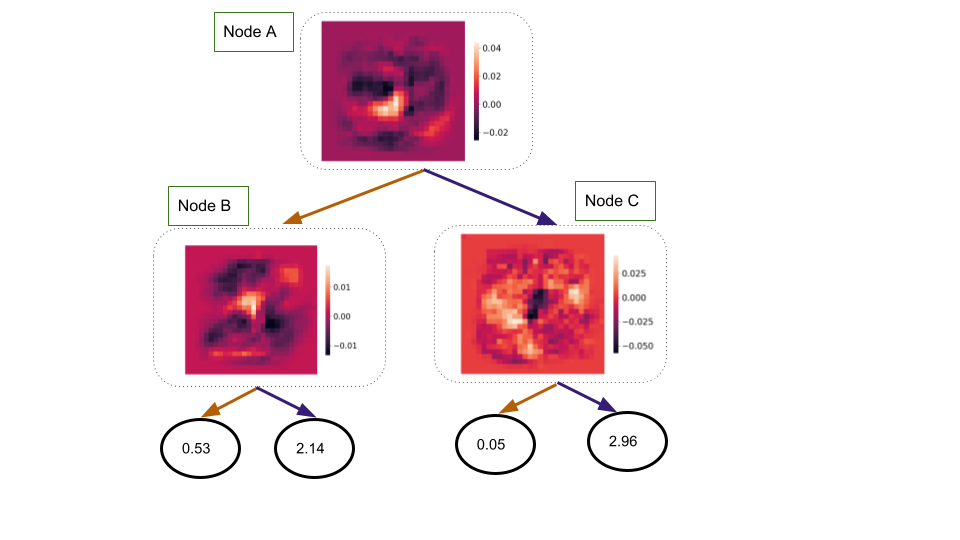}
         \caption{Reward Tree trained using preferences}
         \label{fig:basic small ddt}
     \end{subfigure}
     
     \begin{subfigure}[t]{0.6\linewidth}
         \centering
         \includegraphics[width=\textwidth]{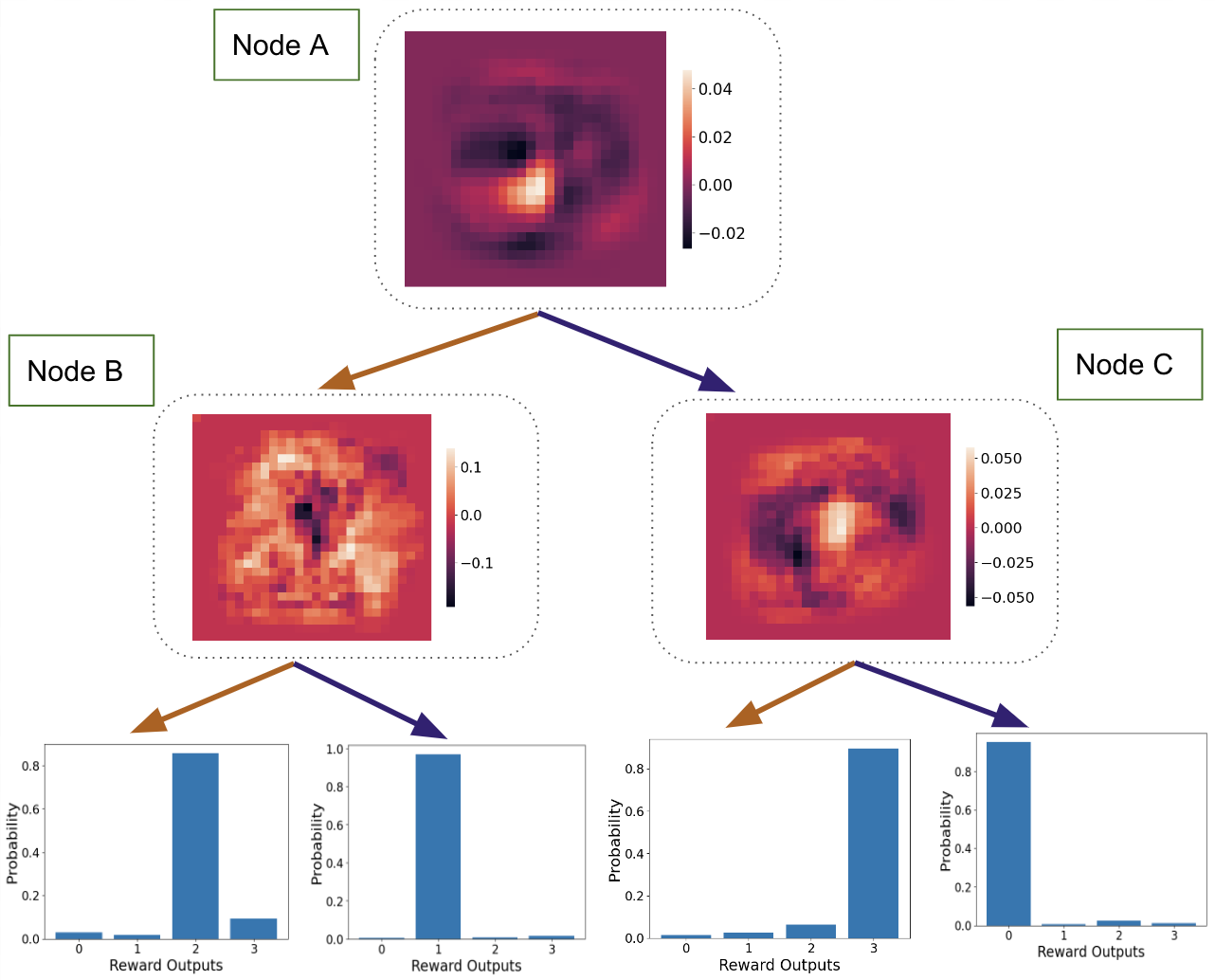}
         \caption{Classification Tree trained on ground truth label}
         \label{fig:classification tree}
     \end{subfigure}
     \hfill
        \caption{\textbf{Visualization of MNSIT (0-3) Reward vs Classification Tree}}
        \label{fig:Reward vs Classification Tree}
\end{figure*}

 We train a DDT with explicit reward labels and a classification loss, as in, we re-produce the classification DDT from [19] and  compare it to reward tree  learned using preferences(refer to Sec 4.2.2 of main paper).

 For comparison of reward tree against the classification tree trained using ground truth labels, we plot the heatmaps of internal nodes in both the trees and our results in Figure~\ref{fig:Reward vs Classification Tree} give evidence that reward tree can capture visual features without any loss in interpretability when compared to the one learnt from simple ground truth labels, even though preferences used here are weaker supervision than ground truth label since preferences used in our experiments are binary as compared to ground truth labels which are 0,1,2,3 corresponding to each actual digit image. This is particularly important in cases where explicit  labels are either missing or are hard to be specified or require intensive user-input efforts.

In Figure~\ref{fig:classification tree}
 Node A activates strongly for pixels in the middle of 1s and 2s, routing them left, while and 0s and 3s are routed right. Node B routes left for vertical pixels in the center and sends 2's left and 1's right (note the light shadow looks like a 2 while the darker shadow  in the middle that looks like a 1). Node C learns to distinguish between 0s and 3s, routing 3s left and 0s right. This is comparable to the activation heatmaps of the node probability distribution at each of the internal node described for reward tree(in Sec 4.2.2 of main paper).

\begin{figure*}[h!]
     \centering
     \begin{subfigure}{0.4\linewidth}
         \centering
         \includegraphics[width=\textwidth]{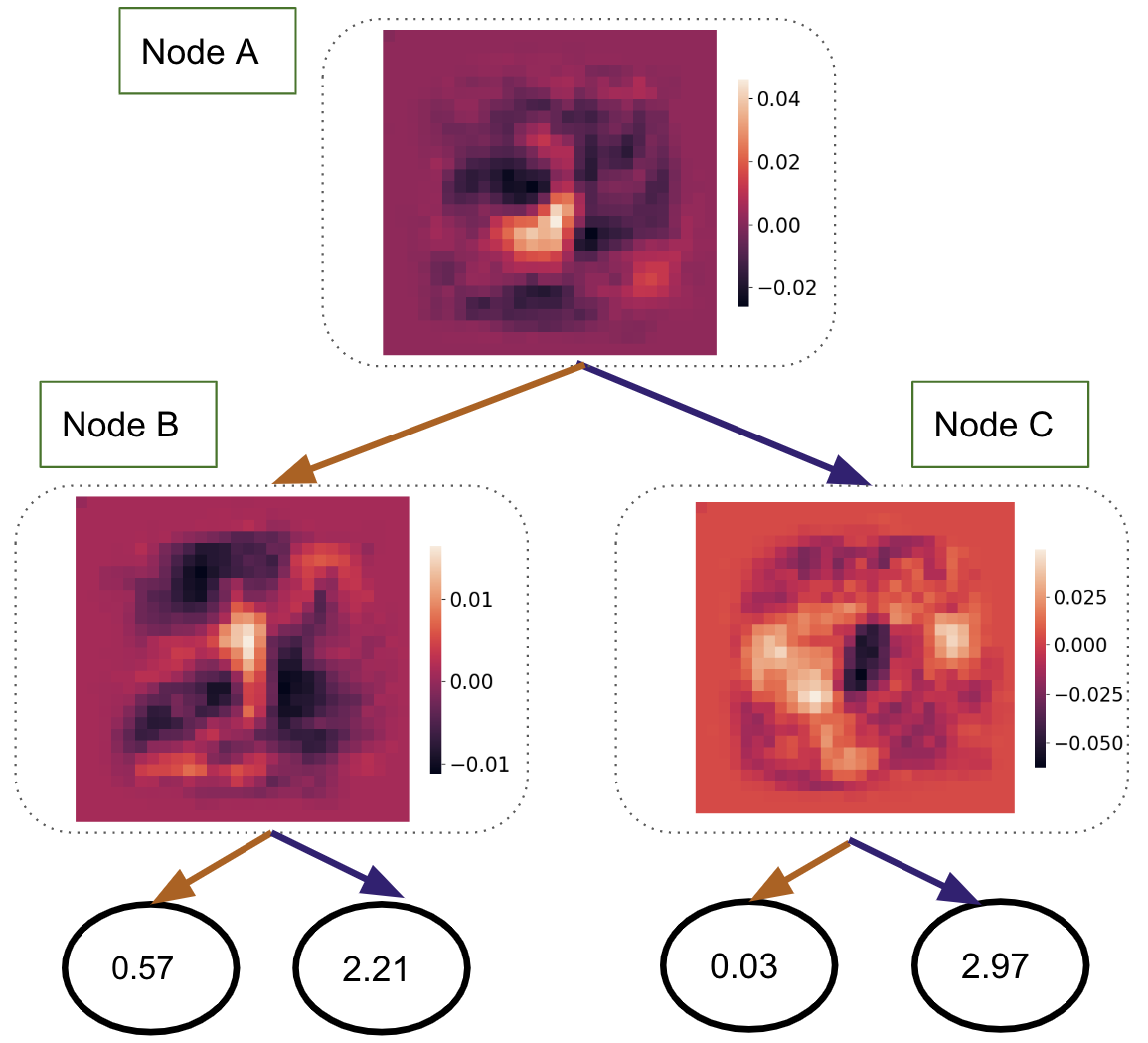}
         \caption{Min-Max Reward Interpolation Leaf DDT}
         \label{fig:basic small ddt}
     \end{subfigure}
     \hfill
     \begin{subfigure}{0.55\linewidth}
         \centering
         \includegraphics[width=\textwidth]{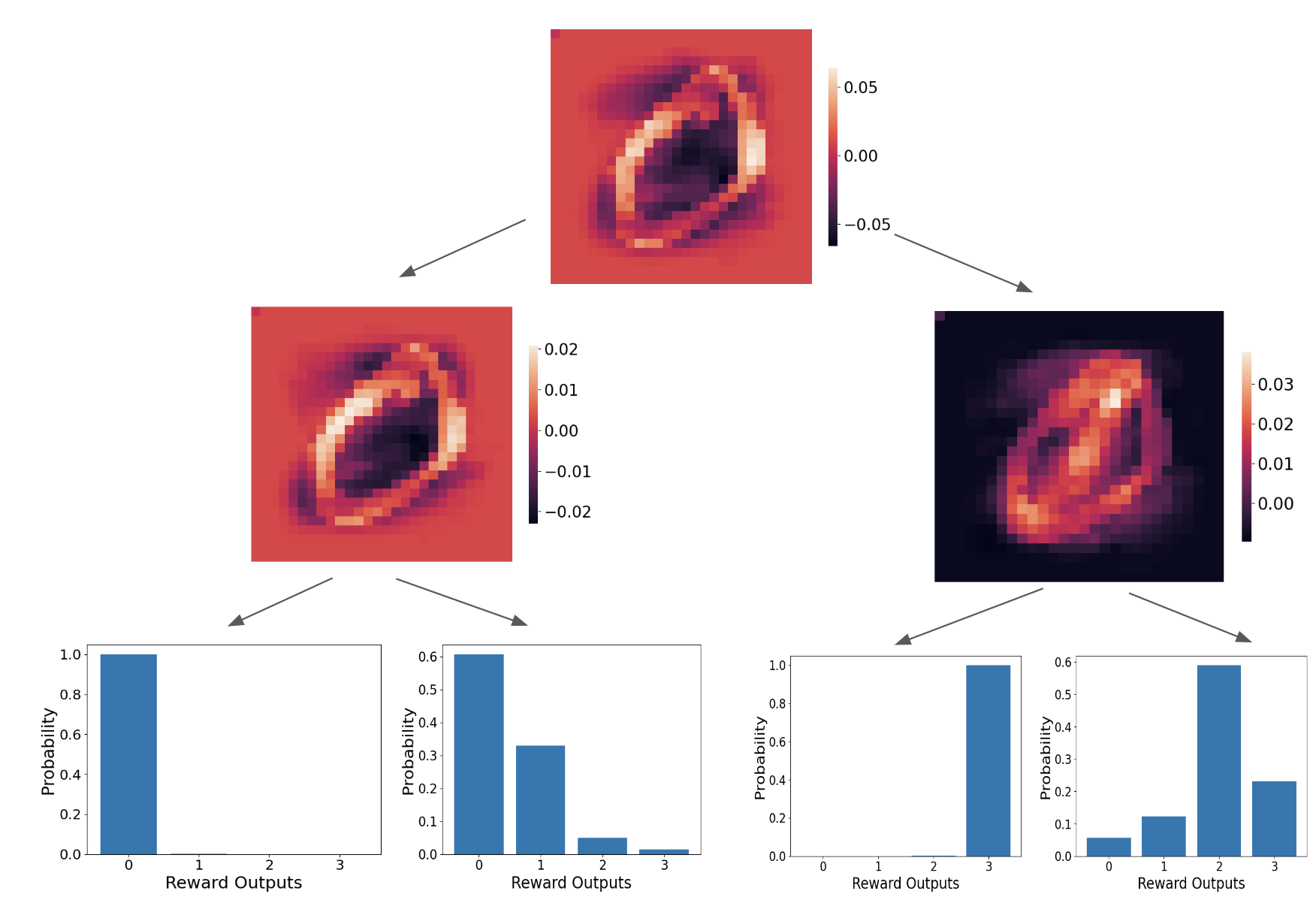}
         \caption{Multi-Class Leaf Reward DDT}
         \label{fig:failed multi-class leaf}
     \end{subfigure}
     \hfill
        \caption{\textbf{Visualization of MNSIT (0-3) Reward Trees: Min-Max Reward Interpolation Leaf vs Multi-Class Leaf}}
        \label{fig:method1 vs 2}
\end{figure*}
\subsection{Min-Max Reward Interpolation Leaf DDT vs Multi-Class  Reward Leaf DDT }
We train and compare two  reward DDTs with simple internal node architecture but with different leaf formulations using the same Bradley-Terry loss over preference demonstration in Figure~\ref{fig:method1 vs 2} by visualizing the activation heatmaps of routing probability distributions for the internal nodes and the leaf distribution for each leaf node. 

In Figure ~\ref{fig:failed multi-class leaf}, each internal node learns to capture almost the same visual feature while the leaf nodes fail to specialize as the argmax output from first two leaf nodes is always a 0 and last two leaf nodes always return a 3.
Multi-class Leaf DDT fails to pick up on individual digit in the trajectory, despite requiring the user to input discrete reward vector whereas in the Min-Max Interpolation Leaf DDT each internal node captures different visual attributes and each of the leaf nodes in the interpolated reward DDT is specialized, even though no discrete reward values were given as an input. 

This shows that Min-Max Reward Interpolation Leaf DDT is beneficial over Multi-Class Reward Leaf DDT with respect to interpretability and also in terms of human-input efforts.
for all states in the pairwise demonstrations.

\subsection { Min-Max Reward Interpolation DDTs with Simple Internal Nodes vs Sophisticated Internal Node }\label{app:comparison_leaf_nodes}
We  compare our 2 methods of constructing internal nodes for a reward DDTs. 

\begin{figure*}[h!]
     \centering
     \begin{subfigure}[t]{0.45\linewidth}
         \centering
         \includegraphics[width=\textwidth]{figures/Basic_Small_DDT.png}
         \caption{Min-Max Reward Interpolation Leaf DDT with Simple Internal Nodes}
         \label{fig:basic small ddt}
     \end{subfigure}
     \hfill
     \begin{subfigure}[t]{0.52\linewidth}
         \centering
         \includegraphics[width=\textwidth]{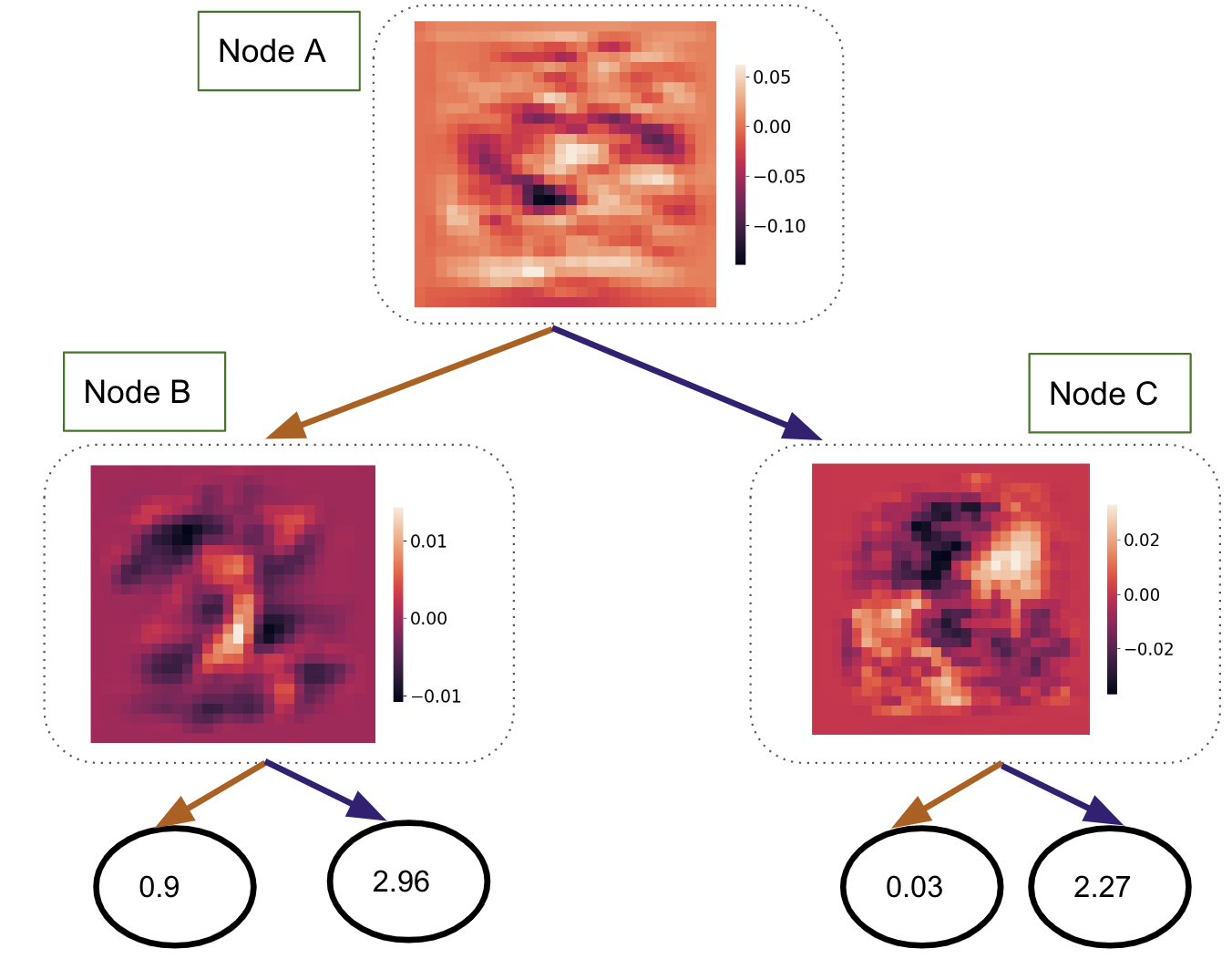}
         \caption{Min-Max Reward Interpolation Leaf DDT with Sophisticated Internal Nodes}
         \label{fig:sophisticated node}
     \end{subfigure}
     \hfill
        \caption{\textbf{Visualization of MNSIT (0-3) Reward Trees :Simple Internal Node vs Sophisticated Internal Node}}
        \label{fig:simple vs sophisticated}
\end{figure*}

Since Min-Max Reward Interpolation Leaf DDT outperforms Multi-Class Reward Leaf DDT, hence we train two different Min-Max Reward Interpolation Leaf DDTs, first one with simple internal nodes and second one with sophisticated internal nodes where a sophisticated internal node contains a single convolutional layer with filter of size 3x3 and stride 1 with Leaky ReLU as the non-linearity followed by the fully connected layer.

In Figure~\ref{fig:sophisticated node}
 Node A activates strongly for pixels in the middle of 1s and 3s, routing them left, while and 0s and 3s are routed right. Node B routes left for vertical pixels in the center and sends 1's left and 3's right (note  the darker shadow in the middle that looks like a 3). Node C learns to distinguish between 0s and 2s, routing 0s left and 2s right. This is comparable to the activation heatmaps of the node probability distribution at each of the internal node described for reward tree(in Sec 4.2.2 of main paper).

Our results depict that in a medium-complexity environment with visual inputs, both DDTs yield relatively equal interpretability but with a higher-complexity environment with larger visual input size such as Atari, the reward DDT with sophisticated node should be used as convolution layer with non-linearity are more powerful in terms of processing an input than a simple fully connected layer.

\subsection{Multi-Class Reward Leaf DDT Regularization}
 Since the DDT with Multi-Class Reward Leaves failed to specialize, this lead us to add the penalty term to the Bradley-Terry preference loss for training the Multi-Class Reward Leaf DDT.

 \begin{figure*}[h!]
     \centering
     \begin{subfigure}{0.9\linewidth}
         \centering
         \includegraphics[width=0.8\textwidth]{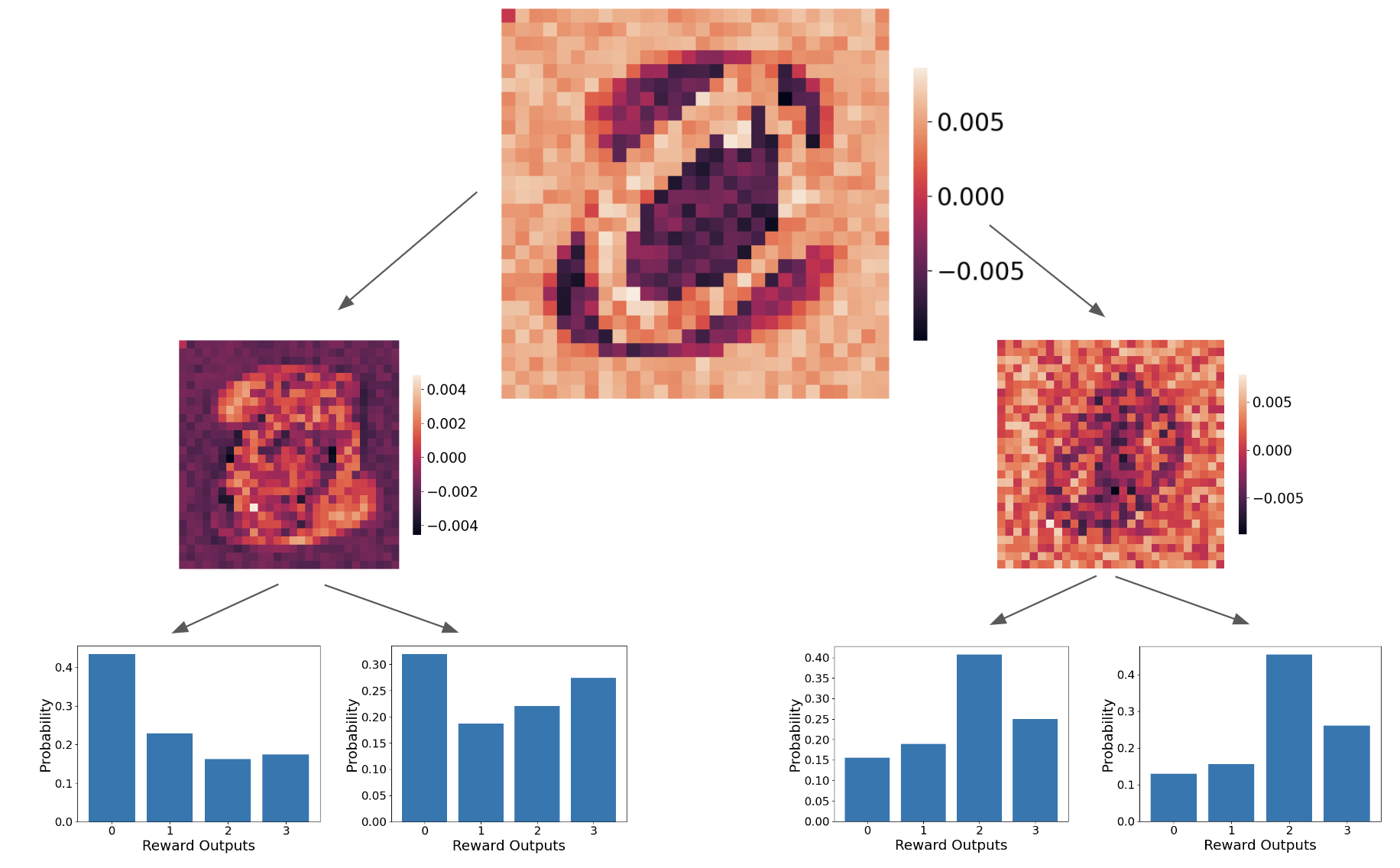}
         \caption{Multi-Class Leaf Reward DDT with penalty calculated over a batch of 50 pairwise preference demonstrations where each demonstration has a single state}
         \label{fig:penalty50-1}
     \end{subfigure}

     \begin{subfigure}{0.9\linewidth}
         \centering
         \includegraphics[width=0.8\textwidth]{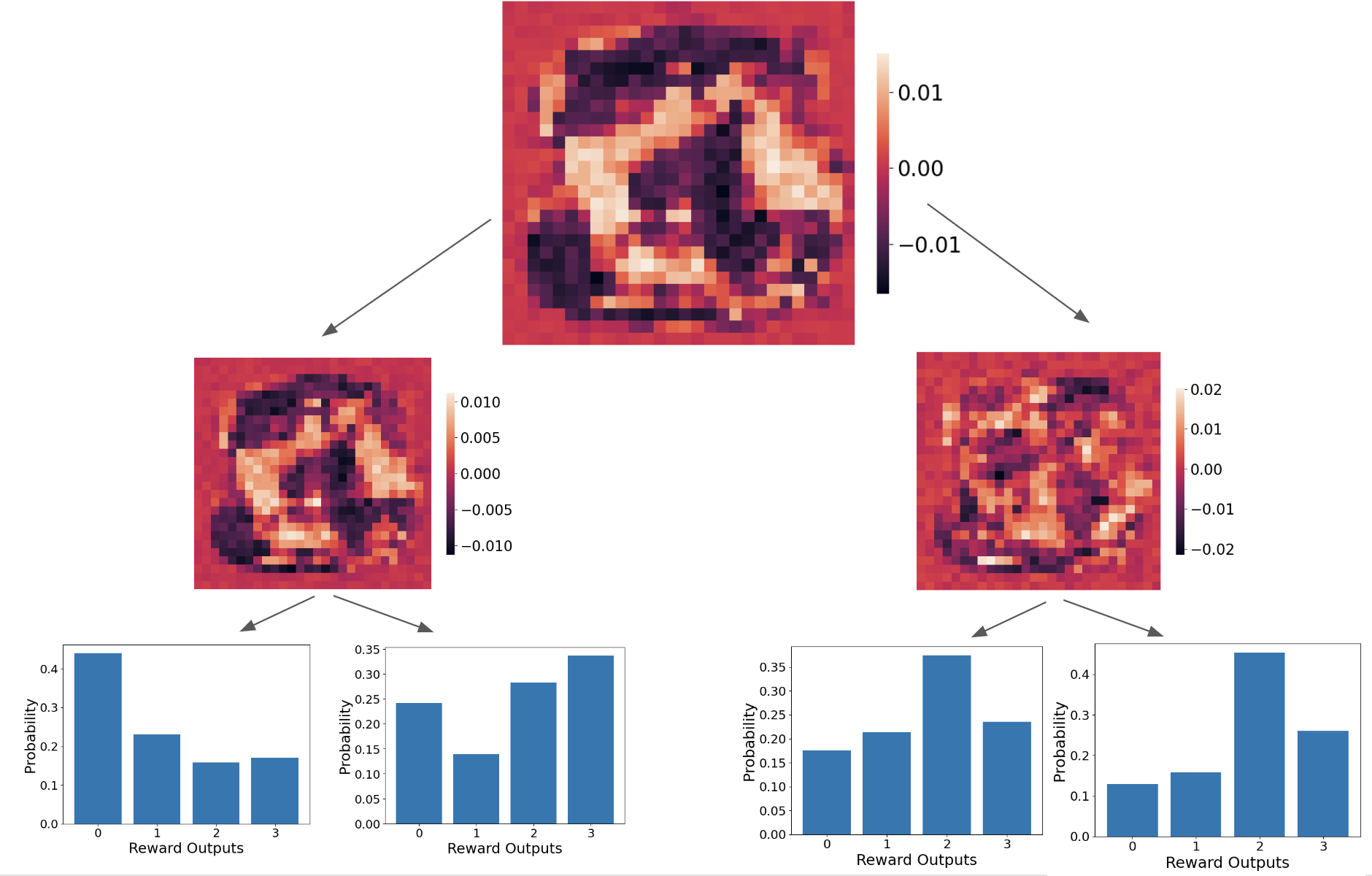}
         \caption{Multi-Class Leaf Reward DDT with penalty calculated over a batch of 50 pairwise preference demonstrations where each demonstration has a single state}
         \label{fig:penalty50-50}
     \end{subfigure}
     
     \hfill
        \caption{Multi-Class Leaf Reward DDT with penalty calculated over different temporal window lengths}
        \label{fig:method1=penalty}
\end{figure*}

 For training the Reward DDT,we calculate penalty over batch of 50 pairwise demonstrations where each demonstration contains a single 28x28 greyscale image.To check interpretability, we plot the activation heatmaps of routing probability distributions for the internal nodes and the leaf distribution for each leaf node in Figure~\ref{fig:penalty50-1} and the resulting plots are hugely pixelated, causing a loss in interpretability. 
 
 Following this, we increase the temporal window size for calculating penalty, as suggested in [19], and thus we calculate penalty over a pair of 50 preference demonstration where each demonstration is now 50 states long, as opposed to previous case where each demonstration contained a single state. 
 And we again visualize  the heatmaps at internal nodes and leaf distributions for each leaf node in Figure ~\ref{fig:penalty50-50}. The heatmaps here are little better in contrast to Figure~\ref{fig:penalty50-1} but still have a huge loss of interpretability as compared to Figure~\ref{fig:failed multi-class leaf}.

{\section{Synthetic trace for MNIST 0-9 Reward DDT}}\label{app:whole_mnist_grid}
\begin{figure}
     \centering
         \includegraphics[width=\textwidth]{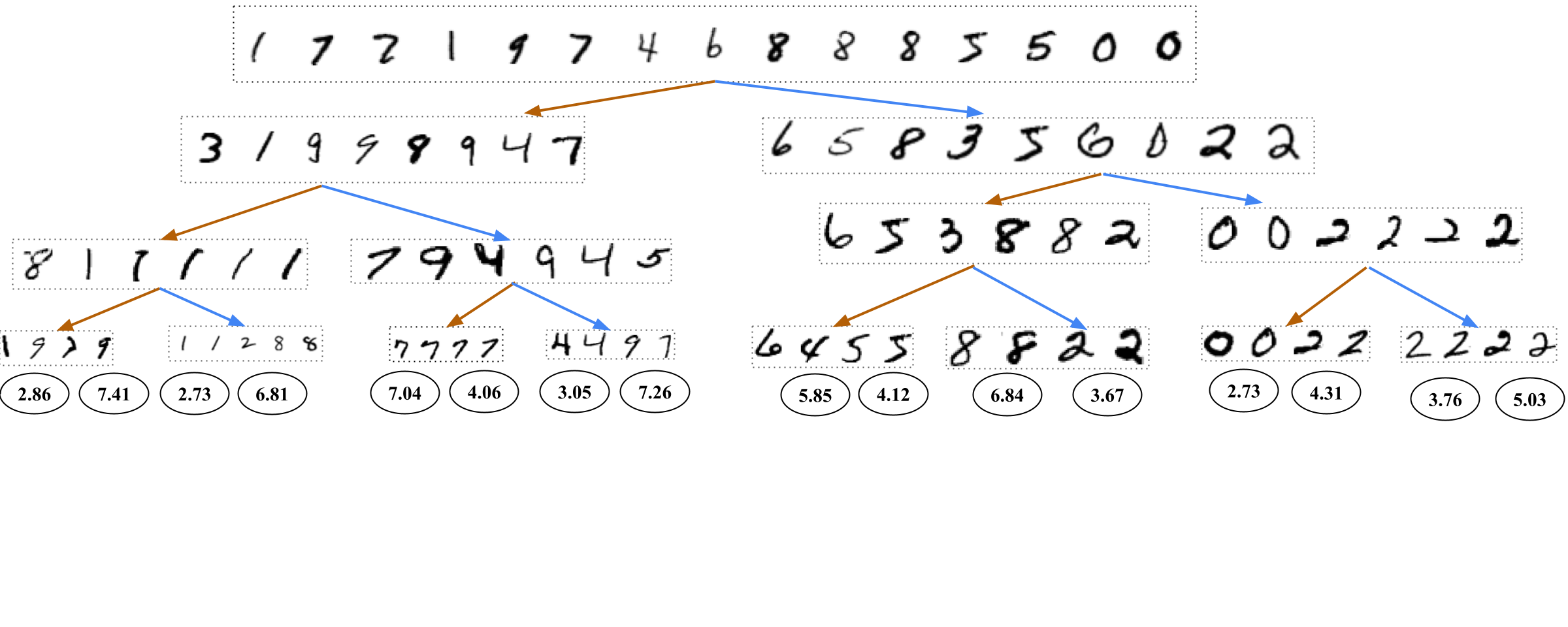}
          \vspace{-2.3cm}
    \caption{Synthetic traces of MNIST 0-9 IL Reward DDT of depth 4.}
    \label{fig:whole mnist traces}
\end{figure}

We create synthetic traces Fig~\ref{fig:whole mnist traces} of learned DDT with IL leaf nodes across all digits in MNIST. 
From the traces, we can observe that root node splits the digits based on whether they have more of vertical formulation or circular formulation. 
The digits with more vertical edges (such as 1,2,3,4,5,8,7,9) are routed to Node B while those with more curved edges (such as 0,2,3,5,6,8) are routed to Node C.
Note some digits such as 2,3,4,8 in the actual MNIST dataset can either be more lean with straight form or can possess more rounded-curve form. The children node of Node B and C then differentiate further between each of the digits routed, as in, children of Node B learn to pick on spread of vertical edges while children of Node C distinguish between forms of curvature. These children's children then learns to pick and specialize in certain specific digits.

\section{Atari} \label{app:atari}
The input to DDT here is a 5-dimensional tensor of size $B\times 2 \times S \times 84 \times 84 \times 4$ where $B$ represents  batch size of pairwise preference demonstrations while 2 is represents of number of demonstrations in a pairwise preference and $S$ represents number of states in a single trajectory. We used batch size $B=25$ and $S=25$.The sophisticated internal node architecture here consists of a single convolution layer with kernel of size $7\times 7$ with a stride of $2$ and LeakyRelu as the non-linearity followed by the fully connected
linear layer for producing the routing probability inside a tree.We used  IL leaf nodes with $R_{\min} =0$ and $R_{\max}=1$. Note that we choose these min and max values for simplicity; though the actual numerical value of $R_{\min}$ and $R_{\max}$ can be chosen at the discretion of the user since policies are invariant to positive scaling and affine.
The baseline T-REX, that we compare to has an architecture similar to ~\citet{christiano2017deep} and consists of 4 convolutional layers of sizes 7x7, 5x5, 3x3 and 3x3 with strides 3,2,1 and 1 respectively, where each convolutional layer has 16 filters and LeakyReLU as non-linearity, followed by a fully connected layer with 64 hidden units and a single scalar output.

\end{document}